\documentclass[twocolumn]{article}

\usepackage{arxiv}%
\usepackage{graphicx}%
\usepackage{multirow}%
\usepackage{amsmath,amssymb,amsfonts}%
\usepackage{amsthm}%
\usepackage{mathrsfs}%
\usepackage{xcolor}%
\usepackage{textcomp}%
\usepackage{manyfoot}%
\usepackage{booktabs}%
\usepackage{algorithm}%
\usepackage{algorithmicx}%
\usepackage{algpseudocode}%
\usepackage{listings}%

\usepackage{comment}
\usepackage{makecell}
\usepackage{hyperref}
\usepackage{float}
\usepackage{subcaption}
\usepackage{lineno}

\usepackage{chngcntr}
\usepackage{natbib}
\usepackage{makecell}

\theoremstyle{plain}
\newtheorem{theorem}{Theorem}[section]
\newtheorem{proposition}[theorem]{Proposition}

\theoremstyle{definition}
\newtheorem{definition}[theorem]{Definition}

\theoremstyle{remark}
\newtheorem{remark}[theorem]{Remark}

\newcommand{\set}[1]{\ensuremath \mathbf{#1}} 

\DeclareMathOperator*{\argmin}{arg\,min\,}
\DeclareMathOperator{\coefsaffine}{\set{A}}
\DeclareMathOperator{\interaffine}{\set{b}}
\DeclareMathOperator{\mvmean}{\boldsymbol{\mu}}
\DeclareMathOperator{\covmat}{\boldsymbol{\Sigma}}

\counterwithin{theorem}{section}

\begin{document}

\title{Optimal transport group counterfactual explanations}



\author{
 Enrique Valero-Leal \\
  Departamento de Inteligencia Artificial\\
  Universidad Politécnica de Madrid \\
  \texttt{enrique.valero@upm.es} \\
   \And
    Bernd Bischl \\
  Department of Statistics, LMU Munich, Germany \\
  Munich Center for Machine Learning (MCML), Germany \\
  \texttt{bernd.bischl@stat.uni-muenchen.de} \\
  \And
 Pedro Larrañaga \\
  Departamento de Inteligencia Artificial\\
  Universidad Politécnica de Madrid \\
  \texttt{pedro.larranaga@fi.upm.es} \\
  \And
 Concha Bielza \\
  Departamento de Inteligencia Artificial\\
  Universidad Politécnica de Madrid \\
  \texttt{mcbielza@fi.upm.es} \\
  \And
    Giuseppe Casalicchio \\
  Department of Statistics, LMU Munich, Germany \\
  Munich Center for Machine Learning (MCML), Germany \\
  \texttt{giuseppe.casalicchio@stat.uni-muenchen.de} \\
}

\twocolumn[
\maketitle
\vspace{1em}
]

\begin{abstract}
    Group counterfactual explanations find a set of counterfactual instances to explain a group of input instances contrastively. 
    However, existing methods either (i) optimize counterfactuals only for a fixed group and do not generalize to new group members, (ii) strictly rely on strong model assumptions (e.g., linearity) for tractability or/and (iii) poorly control the counterfactual group geometry distortion. We instead learn an explicit optimal transport map that sends any group instance to its counterfactual without re-optimization, minimizing the group's total transport cost.
    This enables generalization with fewer parameters, making it easier to interpret the common actionable recourse.
    For linear classifiers, we prove that functions representing group counterfactuals are derived via mathematical optimization, identifying the underlying convex optimization type (QP, QCQP, ...). Experiments show that they accurately generalize,  preserve group geometry and incur only negligible additional transport cost compared to baseline methods. 
    If model linearity cannot be exploited, our approach also significantly outperforms the baselines.
\end{abstract}



\section{Introduction}

\begin{figure*}[ht]
    \centering
    \begin{subfigure}[t]{0.33\textwidth}
        \includegraphics[width=\textwidth]{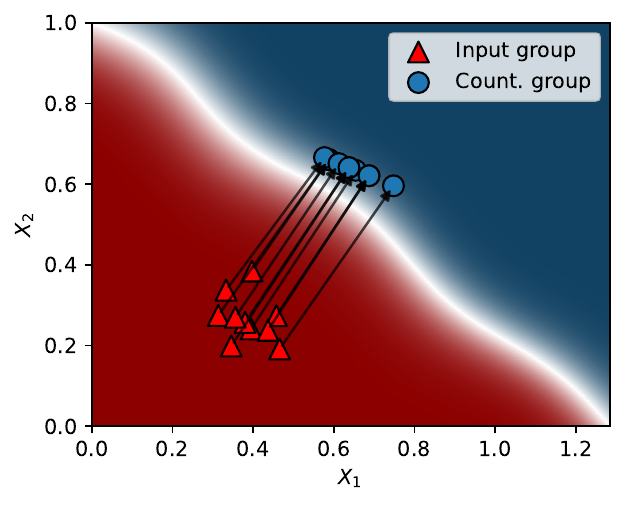}
        \caption{Existing approaches.}
        \label{fig:abstract:a}
    \end{subfigure}
    \hfill
    \begin{subfigure}[t]{0.33\textwidth}
        \includegraphics[width=\textwidth]{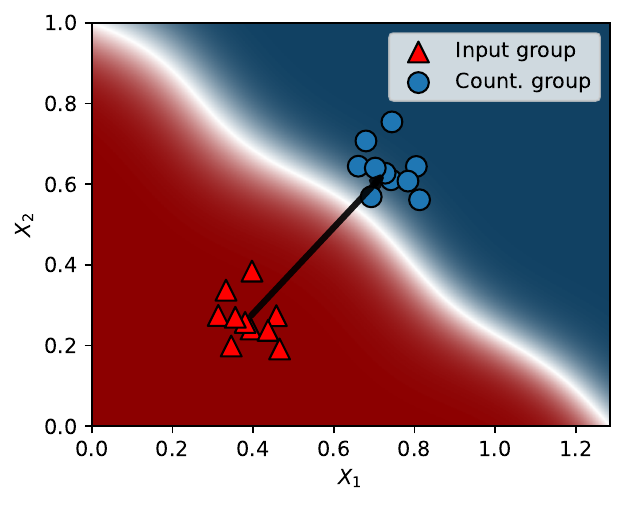}
        \caption{Our proposal.}
        \label{fig:abstract:b}
    \end{subfigure}
    \hfill
    \begin{subfigure}[t]{0.33\textwidth}
        \includegraphics[width=\textwidth]{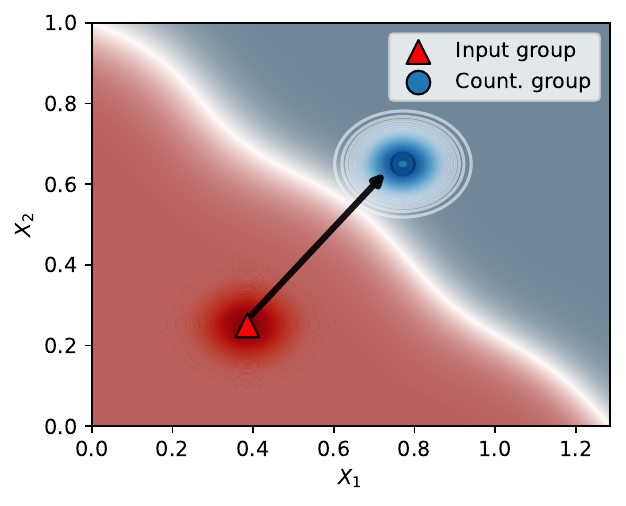}
        \caption{Proposal based on densities.}
        \label{fig:abstract:c}
    \end{subfigure}
    \caption{(a) Current solutions optimize instances individually. (b) Our approach learns a single recourse function for the subgroup, modeling geometric distortion via a bi-Lipschitz parameter $K$. (c) A proposed density-based approach that yields closed-form costs under specific parameterizations (see Section \ref{sec:gaussian_gcfx}).}
    \label{fig:abstract}
\end{figure*}
Explainable AI (XAI) aims to make machine learning models interpretable and trustworthy by providing human-understandable reasons for their predictions \citep{gunning_xaiexplainable_2019, molnar_interpretable_2025}. Counterfactual explanations \citep{wachter_counterfactual_2017, guidotti_counterfactual_2024} have emerged as a prominent approach, describing how minimal changes to input features can alter the prediction of a model.

Group counterfactual explanations \citep{carrizosa_generating_2024, warren_explaining_2024} extend the recourse paradigm to multiple instances simultaneously. Unlike isolated local explanations, this approach aims to ensure similar counterfactuals for similar instances. This capability is critical in scenarios where individual decisions interact with collective constraints. Consider the following motivating examples:

\begin{itemize}
    \item \textbf{Segmented marketing:} In a housing market, a group of similar buyers classified as ``ineligible'' for a desired tier (e.g., luxury offers) may seek recourse that consistently moves them into an eligible tier (e.g., affordable offers).
    
    \item \textbf{Legal policy:} In group litigation (see ``Dieselgate'' \citep{rattalma_dieselgate_2017} for a recent mass-tort case), plaintiffs facing a common eligibility rule for relief may seek group recourse that applies consistent criteria across cases, avoiding case-by-case tailoring.
\end{itemize}

However, the current state of the art \citep{carrizosa_mathematical_2024} is not designed to fully realize this vision: It formulates the problem as a \textbf{point-wise optimization}, finding the optimal counterfactual for each member individually, with only minor restrictions to avoid distortion of the original group characteristics. This can lead to critical failures in the scenarios described above:

\begin{itemize}
    \item \textbf{Violation of feasibility}: For the segmented marketing example, point-wise minimization blindly directs every buyer to the single nearest affordable house. This causes \textit{congestion}, i.e., not enough counterfactual properties to serve the entire group. 
    
    \item \textbf{Violation of fairness}: In the group litigation case, point-wise optimization tailors compensation to individual data noise. This results in a heterogeneous, potentially unfair policy that is difficult to present to a jury, risking rejection.

    \item \textbf{Lack of generalization}: In both examples, if new buyers or plaintiffs appear, existing algorithms need to repeat the optimization process to include them, which is computationally expensive.

    \item \textbf{Large number of parameters}, which may render the problem intractable if mathematical optimization is not possible (non-convex or non-linear classifiers).
\end{itemize}

To address these limitations, we propose a shift from optimizing discrete points to \textit{optimal transport (OT) maps}. By learning a function that represents the counterfactual change 
$\hat{g}: \mathcal X \rightarrow \mathcal X$ (where $\mathcal X$ refers to the feature domain) and that minimizes the transport cost of the entire group, we ensure a better handling of the geometric distortion (i.e., how it contracts, expands, and deforms the input group's geometry), and drastically reduces the dimensionality of the optimization problem, which ensures greater efficiency.

The main contributions of this work are:
\begin{itemize}
    \item A \textbf{theoretical framework} to tackle the group counterfactual problem through OT, which solves the aforementioned problems and allows for user control of the distortion (Section~\ref{sec:methods}). This is further expanded in the appendix with original propositions and definitions.
    \item The \textbf{proposal of different parameterizations} for counterfactual OT maps, including a proof that the associated optimization problem is convex for linear (or linearizable) classifiers, which does not hold for the group method from \cite{carrizosa_mathematical_2024} if a bi-Lipschitz constraint is added, summarized in Table~\ref{tab:model_comparison}.
    \item \textbf{Experimentation and comparison} (Section~\ref{sec:experiments}), demonstrates that our proposal significantly outperforms the baseline w/ bi-Lipschitz constraint. We address scenarios in which mathematical optimization is feasible and those in which heuristics should be used.
\end{itemize}

\section{Background and Related Work}

\subsection{Group Counterfactual Explanations}
Global counterfactual models aim to explain all instances counterfactually. 
To achieve this, \cite{rawal_beyond_2020} and \cite{ley_global_2022} partition the feature space using subgroup discovery \citep{atzmueller_subgroup_2015} and generate rule lists representing actionable recourse per subgroup. 
Similarly, \cite{becker_step_2021} iteratively partitions the space into hypercubes of equally-labeled instances and finds counterfactuals traversing adjacent regions.

In contrast, in group counterfactual approaches, the group is user-defined. \cite{artelt_two-stage_2024} first cluster instances within the group to be explained, and finds different actionable recourses for each cluster. In contrast, \cite{warren_explaining_2024} aim to find counterfactuals with a common actionable recourse by first obtaining individual counterfactuals and then using a post-hoc procedure. \cite{bunay-guisnan_group_2025} add a prior clustering procedure to automatically induce interesting groups, which we also do in this work. 
We do not offer a direct comparison with the aforementioned methods, as they are an addition to classical counterfactual algorithms rather than fully novel proposals.

Other related work includes a unifying view of global, group and local counterfactual explanations \citep{furman_unifying_2025}. \cite{fragkathoulas_facegroup_2026} shift its focus to actionable group counterfactuals, extending the FACE algorithm \citep{poyiadzi_face_2020}. 


\subsection{Optimization of Group Counterfactuals}
\label{sec:carrizosa}

\cite{carrizosa_mathematical_2024} formalize group counterfactuals for linear (or linearizable) classifiers using mathematical optimization. We build on their one-to-one allocation scheme:
\begin{definition}[Group counterfactual (one-to-one)]
\label{def:gcfx}
    Let $\underline{\set{x}} = \{ \set{x}^{(1)},..., \set{x}^{(n)} \}$ with $\set{x}^{(i)} \in \mathcal X = \mathbb R^d$ define a group of $n$ instances. Let $\hat{s}: \mathcal X\times \mathcal C \rightarrow \mathbb R^+$ assign a score to the desired counterfactual label $c' \in \mathcal C$ for instance $\set{x}$, with threshold $\alpha > 0$ and where $\mathcal C$ represents the label domain.
    
    The goal is to find a counterfactual group $\underline{\set{x}'} = \{ \set{x}'^{(1)},..., \set{x}'^{(n)} \}$ by solving:
    \begin{equation}
        \min_{\underline{\set{x}'}} \sum^{n}_{i=1} || \set{x}^{(i)} - \set{x}'^{(i)} ||_2^2, \quad \text{s.t. } \hat{s}(\set{x}'^{(i)}, c') > \alpha, \ \forall i.
        \label{eq:carrizosa_opt}
    \end{equation}
\end{definition}
This defines a pairwise mapping with $nd$ parameters. For linear classifiers, this is a convex quadratic program (QP) and can be divided into $n$ independent problems that are equivalent to the ubiquitous Wachter algorithm \citep{wachter_counterfactual_2017} (see method ``Independent'' in Table~\ref{tab:model_comparison}). The authors suggest adding Lipschitz continuity ($K\geq 0$) so similar instances have similar counterfactuals:
\begin{equation}
    || \set{x}'^{(i)} - \set{x}'^{(j)} ||_2 \leq K \cdot || \set{x}^{(i)} - \set{x}^{(j)} ||_2, \quad \forall i\neq j,
    \label{eq:lipschitz_carrizosa}
\end{equation}
\noindent which makes the problem a quadratically constrained QP (QCQP) if the Euclidean norm is used. It also remains convex (see Proposition~\ref{prop:lipschitz_convex} and method ``Group w/ Lipschitz'' in Table~\ref{tab:model_comparison}).

\subsection{Optimal Transport Theory}
\label{sec:ot}
OT studies efficient transformations between probability distributions. Given distributions $P$ and $Q$\footnote{Throughout, $P$ and $Q$ denote probability distributions (measures). In results about density/volume preservation, we assume $P$ and $Q$ have densities $p$ and $q$ w.r.t. Lebesgue measure. However, we use the notation $P(\set{x})$ and $Q(\set{x})$ for pointwise likelihood for uniformity throughout the text.} on domains $\mathcal{X}$ and $\mathcal{Y}$, and a cost function $c: \mathcal{X} \times \mathcal{Y}$, OT finds the transformation minimizing total cost.

The \textbf{Monge formulation} seeks a map $T: \mathcal{X} \rightarrow \mathcal{Y}$:
\begin{equation}
    \label{eq:monge_map}
    \min_T \int_{\mathcal{X}} c(\set{x}, T(\set{x})) \, dP(\set{x}).
\end{equation}
OT satisfies mass preservation:
\begin{equation}
    \label{eq:mass_preservation}
    P(\set{x}) = Q(T(\set{x})) \cdot |\det DT(\set{x})|,
\end{equation}
where $DT(\mathbf{x})$ is the Jacobian of $T$ (using $D$ as operator of differentiation) evaluated at $\mathbf{x}$.

\textbf{Kantorovich formulation.} If deterministic maps are insufficient, \cite{kantorovich_transfer_1942} proposed probabilistic OT plans $\pi \in \Pi(P, Q)$—joint distributions with marginals $P$ and $Q$:
\begin{equation}
    \label{eq:kantor_map}
    \min_{\pi \in \Pi(P, Q)} \int_{\mathcal{X} \times \mathcal{Y}} c(\set x, \set y) \, d\pi(\set x, \set y).
\end{equation}

The \textbf{Wasserstein distance} is a metric for group distance derived from Equation~(\ref{eq:kantor_map}). The $p$-Wasserstein distance for a metric $d$ is:
\begin{equation}
    \label{eq:wass}
    W_p(P, Q) = \left( \min_{\pi \in \Pi(P, Q)} \int d(\set{x}, \set{y})^p \, d\pi \right)^{1/p}.
\end{equation}
We use the Euclidean distance for $d$ and $p=2$ throughout this work. If $P$ and $Q$ are normally distributed such that $P = \mathcal N(\mvmean_P, \covmat_P)$ and $Q = \mathcal N(\mvmean_Q, \covmat_Q)$, the squared $W_2$ can be expressed as:
\begin{equation}
    \begin{aligned}
        W_2(P, Q)^2  
      & = \| \mvmean_P - \mvmean_Q \|_2^2 \\
      & + \operatorname{Tr}\!\left(
          \covmat_P + \covmat_Q
          - 2 \left( \covmat_Q^{1/2}
                     \covmat_P
                     \covmat_Q^{1/2}
            \right)^{1/2}
        \right).
        \label{eq:wass_normal}
    \end{aligned}
\end{equation}
\paragraph{Optimal Transport Counterfactuals.} 
While literature is sparse, there are mentions of OT in the field of counterfactuals. \cite{lara_transport-based_2024} compare counterfactual distributions derived from causal models and how to reach them with OT plans. \cite{you_distributional_2025}, given an (empirical) input distribution and a desired distribution for the counterfactual label, find a counterfactual distribution. However, they operate with finite samples rather than with the parameters of a distribution.



\section{Methodology}
\label{sec:methods}
\subsection{Motivation}
\label{sec:motivation}
While Definition~\ref{def:gcfx} defines an OT as an allocation solvable with QCQP, the application itself is not defined as an explicit function. Thus, it cannot generalize to unseen group members.

In addition, the theoretical guarantees of a QCQP no longer hold if a non-linear(izable) classifier is used. The large number of parameters ($nd$) of a pairwise-defined application can hinder optimization in these scenarios.

Furthermore, we note that the Lipschitz constraint does not theoretically ensure a similar geometry for the counterfactual, as it prevents points from diverging excessively but does not avoid collapse to a single location. Thus, the counterfactual group can still be highly distorted with respect to the input. This can be avoided by enforcing a bi-Lipschitz condition with parameters $K, k \geq 1$:
\begin{equation}
\begin{split}
         & \tfrac{1}{k}\|\set{x}^{(i)}-\set{x}^{(j)}\|_2 \leq \|\set{x}'^{(i)}-\set{x}'^{(j)}\|_2 \leq K\|\set{x}^{(i)}-\set{x}^{(j)}\|_2, \\
     & \forall i,j \in \{1,..,n\}, i\neq j.
    \label{eq:bilipschitz}
\end{split}
\end{equation}
If this restriction is implemented, the problem becomes non-convex (Proposition~\ref{prop:bilip_noncvx} and method ``Group w/ bi-Lipschitz'' in Table~\ref{tab:model_comparison}).

\subsection{OT Counterfactual Maps}
\label{sec:ot_counterfactuals}
Rather than treating $\underline{\set{x}'}$ as decision variables (Definition~\ref{def:gcfx}), we propose optimizing a function $\hat{g}$ that maps each instance $\set{x}^{(i)} \in \underline{\set{x}}$ to a counterfactual (formalized in Definition~\ref{def:functional_gcfx}).

\begin{definition}[Functional group counterfactual]
    \label{def:functional_gcfx}
    Let $\underline{\set{x}}$, $\hat{s}$, $c'$ and $\alpha$ be as defined in Definition~\ref{def:gcfx} and bi-Lipschitz parameters $K,k \geq 1$.
    
    We aim to find a function $\hat{g}: \mathcal X \rightarrow \mathcal X$ (here we only discuss $\mathcal{X} = \mathbb{R}^d$) that defines the counterfactual group as $\underline{\set{x}'}= \hat{g}(\underline{\set{x}}) = \{ \hat{g}(\set{x}^{(1)}),..., \hat{g}(\set{x}^{(n)}) \},\text{ such that } \forall \set{x}^{(i)} \in \underline{\set{x}}$:
    \begin{equation}
        \hat{g} = \argmin_g \frac{1}{n} \sum^{n}_{i=1} || \set{x}^{(i)} - g(\set{x}^{(i)}) ||_2^2, 
        \label{eq:functional_opt}
    \end{equation}
    \begin{equation}
        \text{s.t. } \hat{s}(g(\set{x}^{(i)}), c') > \alpha,
        \label{eq:functional_constraint}
    \end{equation}
    and s.t. Equation~(\ref{eq:bilipschitz}).
\end{definition}

Depending on the parameterization selected for $\hat{g}$ and the underlying classifier, the problem of Definition~\ref{def:functional_gcfx} ranges from convex QP to non-convex optimization.

Definition~\ref{def:functional_gcfx} is a Monge map, since minimizing the summation of the pairwise distance between input and projection is equivalent to optimizing the $W_2$ distance, thus finding an OT (see Section~\ref{sec:ot}).

Map $\hat{g}$ is assumed to be deterministic, since for continuous variables such mappings are typically optimal \citep{brenier_polar_1991} If the Monge map is insufficient, we can define a Kantorovich plan (Definition~\ref{def:probabilistic_gcfx}).

\begin{definition}[Probabilistic group counterfactual]
    \label{def:probabilistic_gcfx}
    Consider the same problem setup as in Definition~\ref{def:functional_gcfx}. Let $P$ be the probability distribution of $\underline{\set{x}}$ and let $\mathcal Q(\mathcal X)$ represent the space of all possible distributions over the domain $\mathcal X$.

    We now aim to find a distribution $Q \in \mathcal Q(\mathcal X)$ that defines a Kantorovich map (see Equation~(\ref{eq:kantor_map})). 
    \begin{equation}
        \pi_g = \argmin_{\pi\in\Pi(P,Q), \ Q\in \mathcal{Q}(\mathcal{X})} \int_{\mathcal{X}\times \mathcal{X}} c(\mathbf{x},\mathbf{x}')\,\mathrm{d}\pi(\mathbf{x},\mathbf{x}'),
        \label{eq:probabilistic_opt}
    \end{equation}
        \noindent s.t. $\pi_g$ satisfies constraints in Equation~(\ref{eq:functional_constraint})
\end{definition}


\subsection{Density Preservation in Counterfactual Maps}
The bi-Lipschitz constraint (Equation~(\ref{eq:bilipschitz})) allows modeling the extent to which a map $\hat{g}$ distorts the geometry of an input group $\underline{\set{x}}$, in the sense of isotropy and preservation of density. 

An isotropic map is one where the ratio between input and output is constant. Formally, $\hat{g}$ is isotropic if there exists a constant $I \in \mathbb{R}^+$ such that
\begin{equation}
\label{eq:isotropy}
    I = \frac{\|\set{x}^{(i)}-\set{x}^{(j)}\|_2}{\|\hat{g}(\set{x}^{(i)})-\hat{g}(\set{x}^{(j)})\|_2}, \ \forall \set{x}^{(i)},\set{x}^{(j)}.
\end{equation}
It can be verified that this holds if the bi-Lipschitz constraint is enforced with $k=K=1$. 

Density preservation means that the input likelihood of a point equals its posterior likelihood, $P(\set{x}) = Q(\hat{g}(\set{x}))$. This holds when a map is both volume and mass preserving. The latter is given in OT theory (Equation~(\ref{eq:mass_preservation})).

A function is volume preserving if the determinant of its Jacobian equals 1 in its entire domain. For the map $\hat{g}$, formally:
\begin{equation}
|\text{det}\ D\hat{g}(\set{x})|=1, \ \forall\set{x}\in\mathcal X.
\label{eq:volume_preservation}
\end{equation}

With perfect volume preservation, substituting in Equation~(\ref{eq:mass_preservation}) yields the formula for density preservation (i.e. $P(\set x) = Q(\hat g (\set x))$. We prove in the following theorem the relation between bi-Lipschitz continuity and density preservation.

\begin{proposition}[Density Preservation in OT Counterfactual Maps]
\label{prop:density_preservation}
Let $\hat{g}: \mathcal X \to \mathcal X$ (with $\mathcal X \subset \mathbb R^d$) be a fully differentiable $(K,k)$-bi-Lipschitz OT map with $K, k \geq 1$, and let $P$ and $Q$ be the input and target probability measures. Then for every $\mathbf{x} \in \mathcal{X}$:
\begin{equation}
\frac{1}{K^d} P(\mathbf{x}) \leq Q(\hat{g}(\mathbf{x})) \leq k^d P(\mathbf{x}).
\label{eq:density_preservation}
\end{equation}
\end{proposition}

\begin{proof}
    Derived from the Jacobian bounds presented in Proposition~\ref{prop:jacobian_bounds}. Adding the density $P(\set{x})$ to all sides and equating the middle term to $Q(\hat g(\set{x}))$ yields the proof.
\end{proof}

We also link volume preservation with bi-Lipschitz continuity in the appendix, Theorem~\ref{prop:volume_distortion}.


\subsection{Parameterizations for the OT Map}
\label{sec:functional_gcfx}
We propose using interpretable functions $\hat{g}$ to analyze the common actionable recourse in a group counterfactual. A summary of all proposed functions and their characteristics can be found in Table~\ref{tab:model_comparison}. 

\begin{table*}[ht]
\centering
\caption{Comparison of convexity, optimization type, closed form (CF) and complexity class, for both the (squared) $W_2$ and the bi-Lipschitz continuity, BL, and the number of parameters to optimize. Above the horizontal lines, the baseline algorithms are presented. Below, our proposals. 
}
\label{tab:model_comparison}
\begin{tabular}{lccccccc}
\hline
\textbf{Method} & \textbf{Convex} & \textbf{Math opt.} & \bfseries\makecell{$\mathbf{W_2}$ CF} & \bfseries\makecell{$\mathbf{W_2 \ O(\cdot)}$} & \bfseries\makecell{BL CF} & \bfseries\makecell{BL $\mathbf{O(\cdot)}$} & \textbf{\# Parameters} \\
\hline
Independent  & $\checkmark$ & QP &  & $n d$ &  & $1$ & $n d$ \\
Group w/ Lipschitz & $\checkmark$ & QCQP &  & $n d$ &  & $n^2 d$ & $n d$ \\
Group w/ bi-Lipschitz &  & QCQP &  & $n d$ &  & $n^2 d$ & $n d$ \\
\hline
PSD affine transform & $\checkmark$ & SDP &  & $n d^2$ & $\checkmark$ & $d^3$ & $\frac{d(d+1)}{2} + d$ \\
Diagonal affine transform & $\checkmark$ & QP &  & $n d$ & $\checkmark$ & $d$ & $2d$ \\
Gaussian & $\checkmark$ & SDP & $\checkmark$ & $d^2$ & $\checkmark$ & $d^2$ & $d(d+1) + d$ \\
Gaussian commutative & $\checkmark$ & QCQP & $\checkmark$ & $d$ & $\checkmark$ & $d$ & $2d$ \\
Gaussian scaled & $\checkmark$ & QP & $\checkmark$ & $d$ & $\checkmark$ & $1$ & $d + 1$ \\
$k$-GMM & $\checkmark$ & SDP & $\checkmark$ & $kd^2$ & $\checkmark$ & $kd^2$ & $k(d(d+1) + d)$ \\
\hline
\end{tabular}
\end{table*}

We first propose parameterizing the function as an affine map $\hat{g}(\set{x}) = \coefsaffine\set{x} + \interaffine$, which reduces the number of parameters from $n d$ (proposal by \cite{carrizosa_mathematical_2024}) to $d^2 + d$, i.e., it does not scale with the group size $n$ anymore, which might be large. 
If we assume a linear classifier and no Lipschitz constraint, this is a QP problem (Proposition~\ref{prop:affine_qp}). If we introduce a Lipschitz or a bi-Lipschitz constraint, the problem changes to a convex or non-convex QCQP (Proposition~\ref{prop:bilip_noncvx_aff}), respectively.

The affine map yields a theoretical bound for the bi-Lipschitz constraint:
\begin{equation}
K \geq
\sigma_{\max}(\coefsaffine) \text{ and } k\geq
\frac{1}{\sigma_{\min}(\coefsaffine)},
\label{eq:bilip_affine}
\end{equation}
\noindent where $\sigma_{\max}(\coefsaffine)$ and $\sigma_{\min}(\coefsaffine)$ refer to the highest and lowest singular value of $\coefsaffine$ (see Proposition~\ref{prop:lipschitz_sigmas}). 

Assuming $\coefsaffine$ is symmetric and positive semidefinite (PSD), the bi-Lipschitz constraint becomes convex, allowing us to solve the problem using semidefinite programming (SDP).

\begin{proposition}
    \label{prop:affine_psd}
     Consider the setup of a functional group counterfactual problem, Definition~\ref{def:functional_gcfx}, with  $\hat{g}(\set{x}) = \coefsaffine \set{x} +\interaffine$. If $\coefsaffine = \coefsaffine^\top \succeq 0$, then the bi-Lipschitz constraint can be codified as the Loewner order
    \begin{equation}
    \frac{1}{k}I_d \preceq \coefsaffine \preceq KI_d, 
    \label{eq:bilip_sdp}
    \end{equation}
    \noindent where $I_d$ is the $d$-sized identity matrix. This is a set of two affine linear matrix inequalities (LMI), which is convex under an SDP formulation.
\end{proposition}
\begin{proof}
    Directly derived from Proposition~\ref{prop:loewner_bilip}.
\end{proof}

The formulation of Proposition~\ref{prop:affine_psd} reduces the number of parameters to $d(d+1)/2 + d$, i.e., the number of parameters needed to represent a lower diagonal matrix and the parameter $\interaffine$, 
while still allowing all translations and a broad class of linear transformations.
The number of parameters and the overall complexity can be further reduced as follows:

\begin{proposition}[Diagonal $\coefsaffine$ Affine Transform]
    \label{prop:affine_diag}
     The functional group counterfactual problem, Definition~\ref{def:functional_gcfx}, with  $\hat{g}(\set{x}) = \coefsaffine \set{x} +\interaffine$ and $\coefsaffine$ diagonal positive, is QP, and the bi-Lipschitz constraint can be codified as
    \begin{equation}
    \frac{1}{k} \leq d_j \leq K, \forall d_j \in diag(\coefsaffine).
    \label{eq:bilip_diagonal}
\end{equation}
\end{proposition}
\begin{proof}
Since the singular values (and eigenvalues) are exactly the diagonal, the proof becomes trivial.
\end{proof}

As seen in Table~\ref{tab:model_comparison}, these transforms reduce the computational burden for the original problem to varying degrees.

\subsection{Closed-Form for the OT Map Optimization}
\label{sec:gaussian_gcfx}
An alternative to parameterizing a function $\hat{g}$ is to consider the input group $\underline{\set{x}}$ as a probability distribution $P$. 
A well-known option is to assume a multivariate Gaussian structure, which yields a closed-form expression for the transport cost in Equation~(\ref{eq:wass_normal}) and has a similar number of parameters to an affine transform. 

\begin{definition}[Gaussian group counterfactual]
\label{def:functional_gcfx_norm}
    Consider the group counterfactual problem, as introduced in Definition~\ref{def:functional_gcfx}. If we assume $\underline{\set{x}}$ follows a Gaussian distribution $P \sim \mathcal N(\mvmean_P, \covmat_P)$, we can aim to find a target counterfactual distribution  $Q \sim \mathcal N(\mvmean_Q, \covmat_Q)$ by finding the arguments $(\mvmean_Q, \covmat_Q$) that minimize Equation~(\ref{eq:wass_normal}).

    This is still a deterministic (Monge) map, since the application from $P$ to $Q$ is defined by an affine transform $\coefsaffine\set{x}+\interaffine$ \citep{knott_optimal_1984} with
\begin{equation}
\begin{aligned}
\coefsaffine = \coefsaffine^\top &= \covmat_P^{-1/2}(\covmat_P^{1/2}\covmat_Q\covmat_P^{1/2})^{1/2}\covmat_P^{-1/2} \\[1em]
\interaffine &= \mvmean_Q - \coefsaffine\mvmean_P.
\label{eq:affine_normal}
\end{aligned}
\end{equation}
It is easy to see that $\coefsaffine$ is PSD (as $\covmat_P$ and $\covmat_Q$ are).
\end{definition}

Optimizing Equation~(\ref{eq:wass_normal}) is non-trivial, as it contains multiple matrix square roots. 
However, the problem can be addressed by adding $\coefsaffine$ as an additional decision variable.

\begin{theorem}
    \label{the:opt_gaussian}
    Consider the same problem setup as in Definition~\ref{def:functional_gcfx_norm}. The parameterization $Q \sim \mathcal N(\mvmean_Q, \covmat_Q)$ that minimizes Equation~(\ref{eq:wass_normal}) while verifying the bi-Lipschitz constraint (Equation~(\ref{eq:bilip_sdp})) and the classification constraint (Equation~(\ref{eq:functional_constraint})) is found optimizing:
    \begin{equation}
          (\mvmean_Q, \covmat_Q, \coefsaffine ) = \argmin_{\mvmean_Q, \covmat_Q, \coefsaffine}  \| \mvmean_P - \mvmean_Q \|_2^2 + \operatorname{Tr}(\covmat_Q - 2A\covmat_P),
          \label{eq:opt_gaussian}
    \end{equation}
    \begin{equation}
        \begin{bmatrix}
    \covmat_P & \coefsaffine \covmat_P \\
    (\coefsaffine \covmat_P)^\top & \covmat_Q
    \end{bmatrix} \succeq 0,
        \quad
    \covmat_Q \succeq 0,
        \quad
    \coefsaffine \succeq 0
    \end{equation}
\end{theorem}
\begin{proof}
$\coefsaffine$ can be introduced as an additional decision variable that is consistent with Equation~(\ref{eq:affine_normal}), as proven in the more general Proposition~\ref{prop:gassian_A_equiv}.
\end{proof}


It is possible to extend the principles from Theorem~\ref{the:opt_gaussian} to Gaussian mixture models (GMMs), yielding a transport plan with $m$ possible affine transforms (see Appendix~\ref{sec:probabilistic_gcfx}).

Making further assumptions allows deriving $\coefsaffine$ (instead of optimizing it), yielding a convex QCQP.

\begin{theorem}
    \label{prop:opt_gaussian_com}
    Consider the problem of  Definition~\ref{def:functional_gcfx_norm}. If $\covmat_P$ and $\covmat_Q$ commute (i.e., $\covmat_P\covmat_Q = \covmat_Q\covmat_P$), let their eigendecompositions be $\covmat_P = U \Lambda_P U^\top$ and $\covmat_Q = U \Lambda_Q U^\top$. Let $\Lambda_Q = \set{S}_Q^2$. We find the OT map optimizing:
    \begin{equation}
          (\mvmean_Q, \set{S}_Q)= \argmin_{\mvmean_Q, \set{S}_Q}  \| \mvmean_P - \mvmean_Q \|_2^2 +\Vert\covmat_P^{1/2}-U \set{S}_Q U^\top\Vert_{2}^2,
          \label{eq:wass_normal_com}
    \end{equation}
    \begin{equation}
    \text{with }\coefsaffine = \coefsaffine^\top = U \sqrt{\frac{S_Q}{\Lambda_P}} U^\top.
    \label{eq:affine_norm_com}
    \end{equation}
\end{theorem}
\begin{proof}
    The formula for the squared $W_2$ for two Gaussians with a commutative covariance matrix is:
    \begin{equation*}
        W_2(P,Q)^2 =\Vert \mvmean_P-\mvmean_Q\Vert_2^2 +\Vert\covmat_P^{1/2}-\covmat_Q^{1/2}\Vert_{2}^2.
    \end{equation*}
    Equation~(\ref{eq:wass_normal_com}) is a rewriting of this formula using Eigendecomposition. Similarly, Equation~(\ref{eq:affine_norm_com}) is an Eigendecomposition of $\covmat_Q^{1/2}\covmat_P^{-1/2}$, which is derived from Equation~(\ref{eq:affine_normal}) if we assume commutativity.
\end{proof}
Finally, an option to drastically reduce the complexity and number of parameters is presented:

\begin{theorem}
    \label{the:opt_gaussian_scaled}
    Consider Definition (\ref{def:functional_gcfx}), if $\covmat_Q = h\covmat_P$, for some scalar $h>0$ then optimization of Gaussian group counterfactual simplifies to optimize
    \begin{equation*}
          (\mvmean_Q, h)= \argmin_{\mvmean_Q, S}  \| \mvmean_P - \mvmean_Q \|_2^2 + (\sqrt{h} -1 )^2 + \operatorname{Tr}(\covmat_P),
    \end{equation*}
    \noindent deriving $\coefsaffine = \sqrt{h}I_d$
\end{theorem}
\begin{proof}
    Proposition~\ref{prop:wass_norm_scaled} justifies the new formula for the Wasserstein distance. $\coefsaffine$ is derived in an analogous way.
\end{proof}
\section{Experiments}
\label{sec:experiments}
\subsection{Experiment design}
\paragraph{Datasets.}
We evaluate group counterfactuals using 15 binary classification datasets with only numerical features from the benchmark by \cite{grinsztajn_why_2022}. 
These provide real-world data with an appropriate trade-off between complexity and interpretability.

\paragraph{Model.}
We train a regularized logistic regression model on 80\% of the data, reserving the remaining 20\% for counterfactual generation to ensure evaluation on unseen data (see Appendix~\ref{app:model_training} for training details).

\paragraph{Groups considered.} 
We define groups via k-medoids clustering on the test set. 
We generate 10 clusters per label $c \in \mathcal C$ (resulting in 20 groups per dataset and a total of 300) and retain a maximum of 200 random instances per cluster. 
This cap ensures computational feasibility for baseline algorithms that scale poorly with group size, enabling a fair comparison with our proposals (which do not scale quadratically with $n$). Similarly, we sample more instances if the group has fewer than 20.

\paragraph{Baselines.}
We compare our proposal against three methods (corresponding to the first rows of Table~\ref{tab:model_comparison}):
(1) \textbf{Independent} \citep{wachter_counterfactual_2017}: A naive baseline finding counterfactuals for each point independently.
(2) \textbf{Group w/ Lipschitz} \citep{carrizosa_mathematical_2024}: The current state of the art, which couples counterfactuals via a Lipschitz constraint.
(3) \textbf{Group w/ bi-Lipschitz}: A stricter variant of the previous algorithm that enforces bi-Lipschitz, serving as a direct comparison to our methods.


\paragraph{Statistical significance.} 
To assess statistical performance differences, we employ the Friedman test \citep{demsar_statistical_2006} with the \cite{bergmann_improvements_1988} post-hoc procedure. 
We provide the associated critical difference diagrams \citep{demsar_statistical_2006} in Appendix~\ref{app:stats}.

\paragraph{Further experimental setup.}
We set the target class $c'$ as the opposite of the group label, requiring a classification probability $\Pr (c' | \set{x}') > 0.8$. We evaluate bi-Lipschitz constants $K=k \in \{1.01, 1.5, 2, 3.5, 5\}$. The main metrics used for performance evaluation are the $W_2$ distance, the distortion (upper and lower bi-Lipschitz bounds) and the validity (see Appendix~\ref{app:metrics} for details).

\subsection{Mathematical Optimization of OT Group Counterfactuals}
\begin{figure}[htbp]
    \centering
    \includegraphics[width=.48\textwidth]{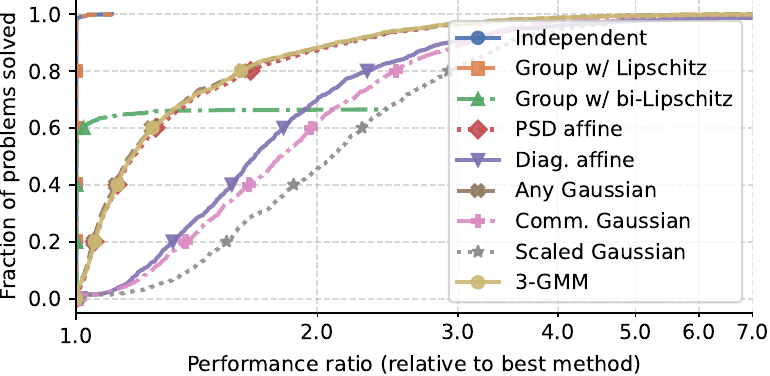}
    \caption{Performance profile for the $W_2$ distance. The color legend is analogous in other figures.}
    \label{fig:math}
\end{figure}

\paragraph{Transport Cost ($W_2$).}
Figure~\ref{fig:math} aggregates the results using performance profiles \citep{dolan_benchmarking_2002}, which indicate what percentage of problems (y-axis) a method can solve within a certain factor of the best solver's metric (x-axis); the higher the curve, the better the solver. The independent and group w/ Lipschitz constraint algorithms achieve minimal $W_2$. The bi-Lipschitz variant maintains efficiency in 65\% of cases but becomes unreliable in upper quantiles due to non-convergence at high $K$ ($K = 1.01$ and $K=1.5$). The gap between the baselines and our proposal narrows for bigger $K$ (see Appendix~\ref{app:k_sensitivity} for further details).

\begin{figure}[htbp]
    \centering
    \includegraphics[width=.48\textwidth]{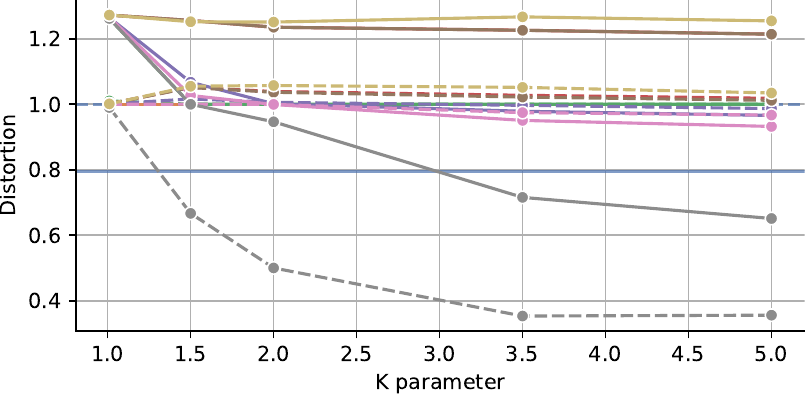}
    \caption{Empirical lower and upper bi-Lipschitz bounds, marked by solid and dashed lines respectively.}
    \label{fig:math_bounds}
\end{figure}

\begin{figure}[htbp]
    \centering
    \includegraphics[width=.48\textwidth]{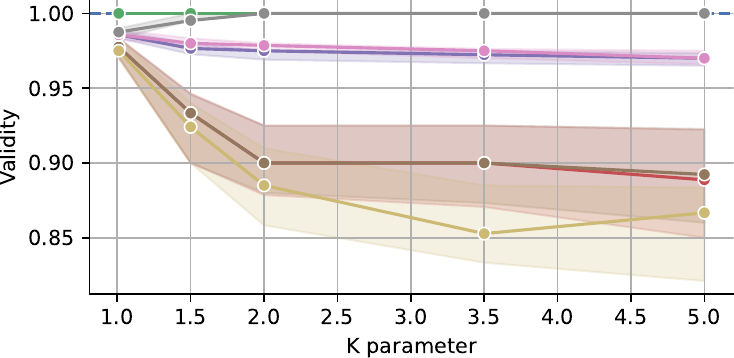}
    \caption{Counterfactual validity lineplot.}
    \label{fig:validity}
\end{figure}

In contrast, our dense maps (PSD affine, Gaussian) cover 90\% of the experiments within a factor lower than 1.7 of the optimal $W_2$. The parameter $K$ is an important factor: With $K=1.01$, our dense proposals find significantly closer group counterfactuals than the bi-Lipschitz baseline. With $K=5$, the opposite is true, and the baseline is able to converge and surpass our proposals.
Sparser transforms (diagonal affine, commutative Gaussian) perform similarly to each other but slightly worse than baselines (factor of 2.3 for 80\% of cases). The sparsest model (scaled Gaussian) yields the highest costs but remains more robust than the bi-Lipschitz constrained baseline.
\paragraph{Distortion bounds.}
Figure~\ref{fig:math_bounds} shows the empirical bounds of the bi-Lipschitz constraint, which characterizes the total distortion. Notably, the Lipschitz constraint (upper bound, see Equation~(\ref{eq:bilipschitz})) is largely satisfied naturally, if it is not explicitly enforced, as in the Independent algorithm or for higher values of $K$. Regarding the bi-Lipschitz lower bound, dense transforms cap their empirical bound at roughly 0.85, regardless of the $K$ provided by the user. Sparse transforms are more sensitive to $K$ but cap at approximately 0.6. For the scaled Gaussian counterfactuals, the lower and upper bounds are identical, indicating perfect isotropy preservation, and capping slightly below 0.4.
\paragraph{Validity.}
Figure~\ref{fig:validity} shows that the validity of solutions decreases with the distortion. For the three baseline algorithms, validity is trivially always 1, since the method does not generalize, but rather finds a valid allocation for the input points. It is also higher in the sparser methods, specifically for the scaled Gaussian model.
\paragraph{Multiobjective optimization.}
We also test our proposal using multiobjective evolutionary metaheuristics rather than mathematical optimization, specifically with NSGA-II \citep{deb_fast_2000}.
This is motivated by the desire (1) to evaluate performance if the classifier $\hat{f}$ is non-linear, as the underlying optimization problem becomes much harder, and (2) to add the distortion as an additional objective rather than a constraint, since it can be hard to user-define the desired bi-Lipschitz constraint. 
Hence, the two objectives to optimize are the $W_2$ distance and the distortion.

The results of this experiment show that all of our proposals yield a significantly larger hypervolume \citep{zitzler_multiobjective_1999} than the baselines. 
The top-performing solutions are the sparser ones, with the Gaussian maps achieving better results than their affine counterparts (details and figures related to the experiment in Appendix~\ref{sec:case_heur}).


\subsection{Post-Hoc Interpretability}
\label{sec:bn}

Following the quantitative results on 15 datasets, we provide a qualitative example for interpreting $\hat g$ by analyzing the generated counterfactual group. Despite the simplicity of affine transforms, interpretation remains non-trivial because, without sparsity constraints, the parameter count grows quadratically with the number of features.

We employ Bayesian networks \citep{koller_probabilistic_2009} as surrogate models for functional group counterfactuals. 
They capture the affine structure in a graphical representation with $2d+1$ nodes: input variables, counterfactual values, and an additional node $K$. 
The underlying graph, causal interpretation and the explicit modeling of $K$ are what place the surrogate model above other interpretability techniques.
Details regarding Gaussian Bayesian networks are provided in appendix~\ref{app_bn}.

\begin{figure}[htbp]
    \centering
    \includegraphics[width=0.85\linewidth]{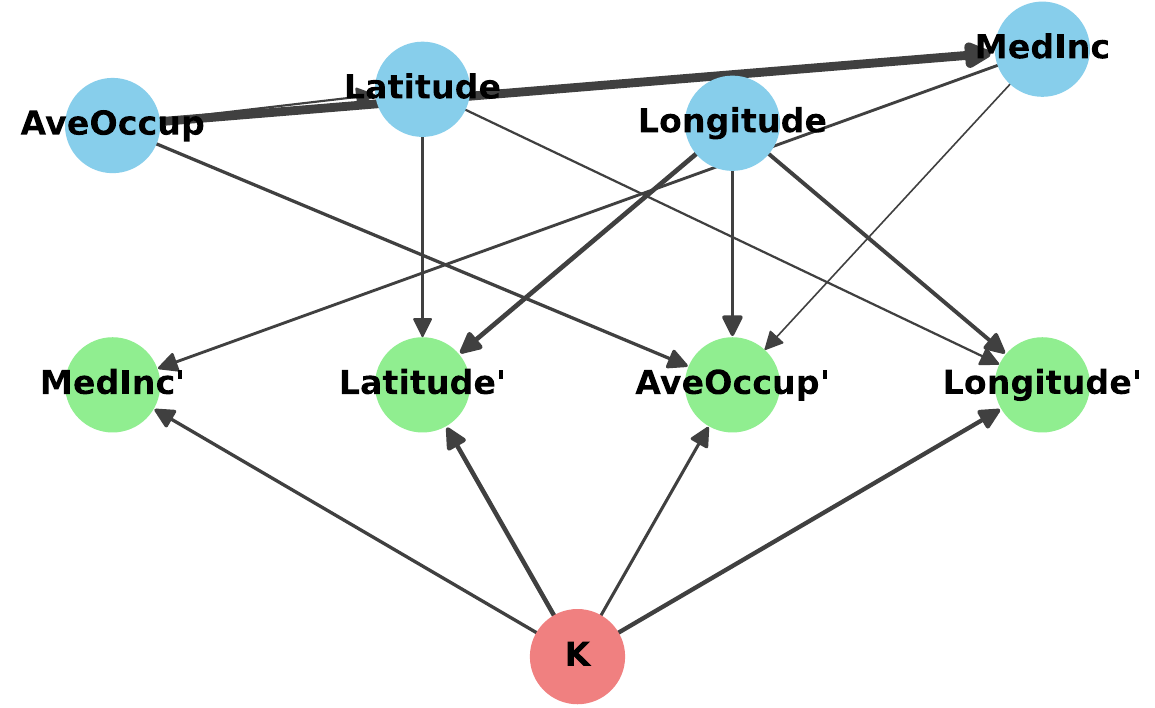}
    \caption{Surrogate Bayesian network. In blue, prior variables. In green, counterfactual variables. $K$ in red. Greater arrow thickness denotes greater effect.}
    \label{fig:bn_california}
\end{figure}
Figure~\ref{fig:bn_california} shows the surrogate Bayesian network for a group of the California dataset (Table~\ref{tab:datasets}) where the houses (instances) are classified as ``cheap'' and we try to flip the classification to ``expensive'', using the PSD affine method. 
We only show the 4 variables that change the most. 
The interdependence between latitude and longitude is evident, as both variables influence their respective counterfactuals. 
In contrast, the counterfactual value of the median income (MedInc) is primarily determined by its original value, although a specific mediation pathway exists from the average occupation variable (AveOccup). 
It is also visible that the effect of $K$ is stronger for latitude.

We can also study the mean regression coefficients, as in Equation~(\ref{eq:bn_coefs}) for the mean of Latitude, $\mu_{Lat.}$. The average pushforward is -0.8, indicating a notable push to the south (data is standardized). Larger values for $K$ (high distortion) reduce this pushforward, almost neutralizing it for $K=5$.
\begin{equation}
    \mu_{Lat.} = -0.8 -0.27Lon. + 0.83Lat. + 0.15K
    \label{eq:bn_coefs}
\end{equation}
\section{Discussion}
\paragraph{Performance and Robustness.}
Our proposals demonstrate high robustness and superiority in the upper quantiles, as the curves rise steadily to cover all the experiments. As such, we conclude that a convex, yet simpler (fewer parameters, see Table~\ref{tab:model_comparison}) representation group counterfactual outperforms a more complete but non-convex alternative.

Additionally, we spot a trade-off between the number of parameters of a map and its performance, i.e., dense representations yield better group counterfactuals.

\paragraph{The Role of bi-Lipschitz Continuity.}
Empirically, the Lipschitz upper bound (dispersion) is often redundant. Neighboring points naturally tend to move in parallel without constraints, unless the local boundary is highly non-linear. This aligns with standard local linearity assumptions (Taylor's theorem) in  machine learning and XAI \citep{bishop_pattern_2006,ribeiro_why_2016}. 

As such, distortion control is done through the lower bound of the bi-Lipschitz constraint (compression), which is not trivially enforced. Strict enforcement in baselines leads to a non-convex problem that fails to converge in approximately $35\%$ of cases for strong constraints ($K=1.01$). Our proposal avoids this instability, finding valid solutions with only a modest cost increase ($30\%$ higher $W_2$ than the unconstrained Wachter baseline) and significantly lower runtime. Empirically, we find that our methods naturally satisfy the lower bound for $K>2.5$.

\paragraph{Validity increases with low distortion.}
 We also observe clear benefits in terms of validity when enforcing isotropy and density preservation, as many points are pushed farther from the decision boundary, ensuring correct classification. 
 In particular, the scaled Gaussian transform ensures a validity of (near to) 1 for all values of $K$. 

\paragraph{Better convergence in metaheuristic optimization.}
Regarding evolutionary multi-objective optimization, our methods outperform all baselines. 
We attribute this to the sparsity of our parameterization. 
While baselines suffer from the curse of dimensionality inherent to heuristic optimization \citep{hooker_testing_1995, bartz-beielstein_experimental_2007}, our reduced search space facilitates faster, better convergence.

\paragraph{Limitations.}
We acknowledge that in the scenarios where the bi-Lipschitz baseline converges (around 65\% of cases), it outperforms all of our proposals. However, this can be explained by the much larger number of parameters (more theoretically powerful) and longer runtime (1000 times slower).

We acknowledge this framework is less useful where congestion or uniformity in actionable recourse are optional; however, this drawback applies to all group counterfactual algorithms, including existing literature.
\section{Conclusions}
We introduce functional and probabilistic group counterfactual explanations using OT maps and show that they can be formalized with fewer parameters than traditional group counterfactual approaches. This implies more simplicity and interpretability. 
We show that the group geometric distortion can be user-tuned via a convex bi-Lipschitz constraint.

Crucially, the convexity of our framework ensures optimization stability, yielding feasible solutions where non-convex baselines fail to converge. These benefits extend to heuristic multiobjective optimization: our proposed methods consistently outperform all baselines.



Future work will explore replacing the direct function $\hat{g}$ with autoregressive processes to capture non-linear paths and model actionability \citep{poyiadzi_face_2020, valero-leal_actionable_2025}. Additionally, while affine maps suffice for tabular data, extending this framework to complex modalities like images will likely require neural network-based OT models \citep{korotin_neural_2023}.

\section*{Acknowledgments}
This work was supported by the Ministry of Education through the University Professor Training (FPU) program fellowship (Enrique Valero-Leal, grant reference FPU21/04812). 

This work was also partially supported by the Ministry of Science, Innovation and Universities under Projects AEI/10.13039/501100011033-PID2022-139977NB-I00 and PLEC2023-010252/MIG-20232016. Also, by the Autonomous Region of Madrid under Project ELLIS Unit Madrid, TEC-2024/COM-89 and IDEA-CM.

The authors gratefully acknowledge the Universidad Politécnica de Madrid (www.upm.es) for providing computing resources on Magerit Supercomputer. 

We also acknowledge the ELLIS PhD program (2021 call, ellis.eu/research/phd-postdoc), as the work was mainly developed on the secondary institution of the PhD student.

\section*{Impact statement}
This work has significant implications for the fairness, feasibility, and scalability of automated decision-making systems.

By defining recourse as an OT map rather than a set of isolated constrained point optimizations, our method ensures a similar actionable recourse for similar individuals. This is crucial for group fairness, particularly in legal and policy contexts (e.g., class-action litigation or credit scoring), where arbitrary disparities in recourse recommendations can undermine public and/or user trust. 


The convex formulation of our method significantly reduces the computational burden required to generate explanations compared to baselines, which are non-convex if we want to preserve the group geometry. Furthermore, the ability to generalize the transport map to unseen data points allows for real-time recourse generation without expensive re-optimization, making XAI more accessible in high-throughput production environments.

Our proposal can also be optimized using evolutionary multiobjective heuristics, which are suitable when dealing with black-box classifiers or when we want to encode the distortion as an objective rather than a constraint.

Although functional counterfactuals promote consistency, we acknowledge that this is not required in every scenario. Additionally, while affine transforms are mathematically interpretable, understanding high-dimensional mappings remains a challenge for lay users, needing further interpretability to prevent misinterpretation of the recommended actions.

\bibliography{references}

@article{grinsztajn_why_2022,
	title = {Why do tree-based models still outperform deep learning on typical tabular data?},
	volume = {35},
	language = {en},
	urldate = {2024-11-13},
	journal = {Advances in Neural Information Processing Systems},
	author = {Grinsztajn, Leo and Oyallon, Edouard and Varoquaux, Gael},
	month = dec,
	year = {2022},
	pages = {507--520},
}

@misc{furman_unifying_2025,
	title = {Unifying perspectives: {Plausible} counterfactual explanations on global, group-wise, and local levels},
	shorttitle = {Unifying perspectives},
	url = {http://arxiv.org/abs/2405.17642},
	doi = {10.48550/arXiv.2405.17642},
	urldate = {2026-01-27},
	publisher = {arXiv},
	author = {Furman, Oleksii and Wielopolski, Patryk and Lenkiewicz, Lukasz and Stefanowski, Jerzy and Zieba, Maciej},
	month = may,
	year = {2025},
	note = {arXiv:2405.17642 [cs]},
}

@inproceedings{fragkathoulas_facegroup_2026,
	address = {Cham},
	title = {{FACEGroup}: {Feasible} and actionable counterfactual explanations for group fairness},
	isbn = {978-3-032-06078-5},
	shorttitle = {{FACEGroup}},
	doi = {10.1007/978-3-032-06078-5_3},
	language = {en},
	booktitle = {Machine {Learning} and {Knowledge} {Discovery} in {Databases}. {Research} {Track}},
	publisher = {Springer Nature Switzerland},
	author = {Fragkathoulas, Christos and Papanikou, Vasiliki and Pitoura, Evaggelia and Terzi, Evimaria},
	editor = {Ribeiro, Rita P. and Pfahringer, Bernhard and Japkowicz, Nathalie and Larrañaga, Pedro and Jorge, Alípio M. and Soares, Carlos and Abreu, Pedro H. and Gama, João},
	year = {2026},
	pages = {41--59},
}

@article{bunay-guisnan_group_2025,
	title = {Group counterfactual explanations: {A} use case to support students at risk of dropping out in online education},
	volume = {15},
	copyright = {http://creativecommons.org/licenses/by/3.0/},
	issn = {2079-9292},
	shorttitle = {Group counterfactual explanations},
	url = {https://www.mdpi.com/2079-9292/15/1/51},
	doi = {10.3390/electronics15010051},
	language = {en},
	number = {1},
	urldate = {2026-01-27},
	journal = {Electronics},
	publisher = {MDPI},
	author = {Buñay-Guisñan, Pamela and Cano, Alberto and Anguera, Aurea and Lara, Juan A. and Romero, Cristóbal},
	month = dec,
	year = {2025},
	pages = {51},
}

@article{zitzler_multiobjective_1999,
	title = {Multiobjective evolutionary algorithms: a comparative case study and the strength {Pareto} approach},
	volume = {3},
	issn = {1941-0026},
	shorttitle = {Multiobjective evolutionary algorithms},
	url = {https://ieeexplore.ieee.org/abstract/document/797969},
	doi = {10.1109/4235.797969},
	number = {4},
	urldate = {2026-01-27},
	journal = {IEEE Transactions on Evolutionary Computation},
	author = {Zitzler, E. and Thiele, L.},
	month = nov,
	year = {1999},
	pages = {257--271},
}

@article{becker_step_2021,
	title = {A step towards global counterfactual explanations: {Approximating} the feature space through hierarchical division and graph search},
	volume = {1},
	issn = {25829793},
	shorttitle = {A step towards global counterfactual explanations},
	url = {https://www.oajaiml.com/uploads/archivepdf/18151107.pdf},
	doi = {10.54364/AAIML.2021.1107},
	language = {en},
	number = {2},
	urldate = {2025-10-23},
	journal = {Advances in Artificial Intelligence and Machine Learning},
	author = {Becker, Maximilian and Burkart, Nadia and Birnstill, Pascal and Beyerer, Jürgen},
	year = {2021},
	pages = {90--110},
}

@inproceedings{ribeiro_why_2016,
	address = {New York, NY, USA},
	title = {Why should {I} trust you?: {Explaining} the predictions of any classifier},
	isbn = {978-1-4503-4232-2},
	shorttitle = {"why should {I} trust you?},
	url = {https://doi.org/10.1145/2939672.2939778},
	doi = {10.1145/2939672.2939778},
	urldate = {2025-12-24},
	booktitle = {Proceedings of the 22nd {ACM} {SIGKDD} {International} {Conference} on {Knowledge} {Discovery} and {Data} {Mining}},
	publisher = {ACM},
	author = {Ribeiro, Marco Tulio and Singh, Sameer and Guestrin, Carlos},
	year = {2016},
	pages = {1135--1144},
}

@book{molnar_interpretable_2025,
	edition = {3rd},
	title = {Interpretable {Machine} {Learning}},
	isbn = {978-3-911578-03-5},
	url = {https://christophm.github.io/interpretable-ml-book},
	author = {Molnar, Christoph},
	year = {2025},
}

@article{diamond_cvxpy_2016,
	title = {{CVXPY}: {A} python-embedded modeling language for convex optimization},
	volume = {17},
	issn = {1532-4435},
	shorttitle = {{CVXPY}},
	url = {https://dl.acm.org/doi/10.5555/2946645.3007036},
	number = {1},
	urldate = {2026-01-13},
	journal = {Journal of Machine Learning Research},
	author = {Diamond, Steven and Boyd, Stephen},
	year = {2016},
	pages = {2909--2913},
}

@article{demsar_statistical_2006,
	title = {Statistical comparisons of classifiers over multiple data sets.},
	volume = {7},
	issn = {1532-4435},
	url = {https://openurl.ebsco.com/contentitem/gcd:20018430?sid=ebsco:plink:crawler&id=ebsco:gcd:20018430},
	language = {es},
	number = {1},
	urldate = {2025-11-05},
	journal = {Journal of Machine Learning Research},
	author = {Demšar, Janez},
	month = jan,
	year = {2006},
	pages = {1--30},
}

@inproceedings{you_distributional_2025,
	title = {Distributional counterfactual explanations with optimal transport},
	url = {https://openreview.net/forum?id=cGnnwEm2gu},
	language = {en},
	urldate = {2025-10-30},
	booktitle = {Proceedings of the 28th {International} {Conference} on {Artificial} {Intelligence} and {Statistics}},
	publisher = {PMLR},
	author = {You, Lei and Cao, Lele and Nilsson, Mattias and Zhao, Bo and Lei, Lei},
	month = feb,
	year = {2025},
	pages = {1135--1143},
}

@inproceedings{warren_explaining_2024,
	title = {Explaining multiple instances counterfactually: {User} tests of group-counterfactuals for {XAI}},
	isbn = {978-3-031-63646-2},
	shorttitle = {Explaining multiple instances counterfactually},
	doi = {10.1007/978-3-031-63646-2_14},
	language = {en},
	booktitle = {Case-{Based} {Reasoning} {Research} and {Development}},
	publisher = {Springer},
	author = {Warren, Greta and Delaney, Eoin and Guéret, Christophe and Keane, Mark T.},
	year = {2024},
	pages = {206--222},
}

@misc{valero-leal_actionable_2025,
	title = {Actionable counterfactual explanations using {Bayesian} networks and path planning with applications to environmental quality improvement},
	url = {http://arxiv.org/abs/2508.02634},
	doi = {10.48550/arXiv.2508.02634},
	urldate = {2025-10-30},
	publisher = {arXiv},
	author = {Valero-Leal, Enrique and Larrañaga, Pedro and Bielza, Concha},
	month = aug,
	year = {2025},
	note = {arXiv:2508.02634},
}

@book{spirtes_causation_2000,
	title = {Causation, {Prediction}, and {Search}},
	publisher = {The MIT Press},
	author = {Spirtes, Peter and Glymour, Clark N and Scheines, Richard and Heckerman, David},
	year = {2000},
}

@book{rattalma_dieselgate_2017,
	title = {The {Dieselgate}: {A} {Legal} {Perspective}},
	isbn = {978-3-319-48323-8},
	shorttitle = {The {Dieselgate}},
	language = {en},
	publisher = {Springer},
	author = {Rattalma, Marco Frigessi},
	month = jan,
	year = {2017},
}

@inproceedings{poyiadzi_face_2020,
	series = {{AIES} '20},
	title = {{FACE}: {Feasible} and actionable counterfactual explanations},
	isbn = {978-1-4503-7110-0},
	shorttitle = {{FACE}},
	url = {https://dl.acm.org/doi/10.1145/3375627.3375850},
	doi = {10.1145/3375627.3375850},
	urldate = {2024-11-13},
	booktitle = {Proceedings of the {AAAI}/{ACM} {Conference} on {AI}, {Ethics}, and {Society}},
	publisher = {ACM},
	author = {Poyiadzi, Rafael and Sokol, Kacper and Santos-Rodriguez, Raul and De Bie, Tijl and Flach, Peter},
	month = feb,
	year = {2020},
	pages = {344--350},
}

@book{koller_probabilistic_2009,
	title = {Probabilistic {Graphical} {Models}: {Principles} and {Techniques}},
	isbn = {978-0-262-01319-2},
	shorttitle = {Probabilistic {Graphical} {Models}},
	language = {en},
	publisher = {The MIT Press},
	author = {Koller, Daphne and Friedman, Nir},
	month = jul,
	year = {2009},
}

@article{guidotti_counterfactual_2024,
	title = {Counterfactual explanations and how to find them: {Literature} review and benchmarking},
	volume = {38},
	issn = {1573-756X},
	shorttitle = {Counterfactual explanations and how to find them},
	url = {https://doi.org/10.1007/s10618-022-00831-6},
	doi = {10.1007/s10618-022-00831-6},
	language = {en},
	number = {5},
	urldate = {2025-07-10},
	journal = {Data Mining and Knowledge Discovery},
	author = {Guidotti, Riccardo},
	year = {2024},
	pages = {2770--2824},
}

@inproceedings{deb_fast_2000,
	title = {A fast elitist non-dominated sorting genetic algorithm for multi-objective optimization: {NSGA}-{II}},
	isbn = {978-3-540-45356-7},
	shorttitle = {A fast elitist non-dominated sorting genetic algorithm for multi-objective optimization},
	doi = {10.1007/3-540-45356-3_83},
	language = {en},
	booktitle = {Parallel {Problem} {Solving} from {Nature} {PPSN} {VI}},
	publisher = {Springer},
	author = {Deb, Kalyanmoy and Agrawal, Samir and Pratap, Amrit and Meyarivan, T.},
	year = {2000},
	pages = {849--858},
}

@inproceedings{dandl_multi-objective_2020,
	title = {Multi-objective counterfactual explanations},
	isbn = {978-3-030-58112-1},
	doi = {10.1007/978-3-030-58112-1_31},
	language = {en},
	booktitle = {Parallel {Problem} {Solving} from {Nature} – {PPSN} {XVI}},
	publisher = {Springer},
	author = {Dandl, Susanne and Molnar, Christoph and Binder, Martin and Bischl, Bernd},
	year = {2020},
	pages = {448--469},
}

@book{bynum_pyomo_2021,
	edition = {Third},
	title = {Pyomo — {Optimization} {Modeling} in {Python}},
	volume = {67},
	copyright = {http://www.springer.com/tdm},
	isbn = {978-3-030-68927-8 978-3-030-68928-5},
	url = {http://link.springer.com/10.1007/978-3-030-68928-5},
	doi = {10.1007/978-3-030-68928-5},
	language = {en},
	urldate = {2026-01-13},
	publisher = {Springer},
	author = {Bynum, Michael L. and Hackebeil, Gabriel A. and Hart, William E. and Laird, Carl D. and Nicholson, Bethany L. and Siirola, John D. and Watson, Jean-Paul and Woodruff, David L.},
	year = {2021},
}

@inproceedings{bergmann_improvements_1988,
	title = {Improvements of general multiple test procedures for redundant systems of hypotheses},
	isbn = {978-3-642-52307-6},
	doi = {10.1007/978-3-642-52307-6_8},
	language = {de},
	booktitle = {Multiple {Hypotheses} {Testing}},
	publisher = {Springer},
	author = {Bergmann, B. and Hommel, G.},
	year = {1988},
	pages = {100--115},
}

@inproceedings{bartz-beielstein_experimental_2007,
	series = {{GECCO} '07},
	title = {Experimental research in evolutionary computation},
	isbn = {978-1-59593-698-1},
	url = {https://dl.acm.org/doi/10.1145/1274000.1274102},
	doi = {10.1145/1274000.1274102},
	urldate = {2025-12-13},
	booktitle = {Proceedings of the 9th {Annual} {Conference} {Companion} on {Genetic} and {Evolutionary} {Computation}},
	publisher = {ACM},
	author = {Bartz-Beielstein, Thomas and Preuss, Mike},
	month = jul,
	year = {2007},
	pages = {3001--3020},
}

@inproceedings{artelt_two-stage_2024,
	title = {A two-stage algorithm for cost-efficient multi-instance counterfactual explanations},
	url = {https://ui.adsabs.harvard.edu/abs/2024arXiv240301221A},
	doi = {10.48550/arXiv.2403.01221},
	urldate = {2025-10-23},
	booktitle = {Joint {Proceedings} of the {XAI} 2024 {Late}-breaking {Work}, {Demos} and {Doctoral} {Consortium} (co-located with the 2nd {World} {Conference} on {eXplainable} {Artificial} {Intelligence})},
	author = {Artelt, André and Gregoriades, Andreas},
	month = mar,
	year = {2024},
	pages = {233--240},
}

@misc{mosek_aps_mosek_2025,
	title = {{MOSEK} optimizer {API} for {Python}. {Version} 11.0.29},
	url = {https://docs.mosek.com/latest/pythonapi/index.html},
	author = {MOSEK Aps},
	year = {2025},
}

@article{agrawal_rewriting_2017,
	title = {A rewriting system for convex optimization problems},
	volume = {5},
	doi = {10.1080/23307706.2017.1397554},
	number = {1},
	journal = {Journal of Control and Decision},
	author = {Agrawal, Akshay and Verschueren, Robin and Diamond, Steven and Boyd, Stephen},
	year = {2017},
	pages = {42--60},
}

@inproceedings{korotin_neural_2023,
	title = {Neural optimal transport},
	url = {https://openreview.net/forum?id=d8CBRlWNkqH},
	language = {en},
	urldate = {2025-11-13},
	booktitle = {Proceedings of the 11th {International} {Conference} on {Learning} {Representations}},
	author = {Korotin, Alexander and Selikhanovych, Daniil and Burnaev, Evgeny},
	year = {2023},
}

@article{hart_pyomo_2011,
	title = {Pyomo: {Modeling} and solving mathematical programs in {Python}},
	volume = {3},
	issn = {1867-2957},
	shorttitle = {Pyomo},
	url = {https://doi.org/10.1007/s12532-011-0026-8},
	doi = {10.1007/s12532-011-0026-8},
	language = {en},
	number = {3},
	urldate = {2026-01-13},
	journal = {Mathematical Programming Computation},
	author = {Hart, William E. and Watson, Jean-Paul and Woodruff, David L.},
	month = sep,
	year = {2011},
	pages = {219--260},
}

@article{blank_pymoo_2020,
	title = {Pymoo: {Multi}-objective optimization in {Python}},
	volume = {8},
	issn = {2169-3536},
	shorttitle = {Pymoo},
	url = {https://ieeexplore.ieee.org/document/9078759},
	doi = {10.1109/ACCESS.2020.2990567},
	urldate = {2026-01-13},
	journal = {IEEE Access},
	author = {Blank, Julian and Deb, Kalyanmoy},
	year = {2020},
	pages = {89497--89509},
}

@book{bishop_pattern_2006,
	title = {Pattern {Recognition} and {Machine} {Learning}},
	url = {https://link.springer.com/book/9780387310732},
	language = {en},
	urldate = {2025-12-24},
	publisher = {Springer},
	author = {Bishop, Christopher},
	year = {2006},
}

@article{dolan_benchmarking_2002,
	title = {Benchmarking optimization software with performance profiles},
	volume = {91},
	issn = {1436-4646},
	url = {https://doi.org/10.1007/s101070100263},
	doi = {10.1007/s101070100263},
	language = {en},
	number = {2},
	urldate = {2026-01-15},
	journal = {Mathematical Programming},
	author = {Dolan, Elizabeth D. and Moré, Jorge J.},
	month = jan,
	year = {2002},
	pages = {201--213},
}

@misc{gurobi_optimization_llc_gurobi_2024,
	title = {Gurobi {Optimizer} {Reference} {Manual}},
	url = {https://www.gurobi.com},
	author = {Gurobi Optimization, LLC},
	year = {2024},
}

@article{hooker_testing_1995,
	title = {Testing heuristics: {We} have it all wrong},
	volume = {1},
	issn = {1572-9397},
	shorttitle = {Testing heuristics},
	url = {https://doi.org/10.1007/BF02430364},
	doi = {10.1007/BF02430364},
	language = {en},
	number = {1},
	urldate = {2025-12-13},
	journal = {Journal of Heuristics},
	author = {Hooker, J. N.},
	month = sep,
	year = {1995},
	pages = {33--42},
}

@article{carrizosa_generating_2024,
	title = {Generating collective counterfactual explanations in score-based classification via mathematical optimization},
	volume = {238},
	issn = {0957-4174},
	url = {https://www.sciencedirect.com/science/article/pii/S0957417423024569},
	doi = {10.1016/j.eswa.2023.121954},
	urldate = {2025-11-24},
	journal = {Expert Systems with Applications},
	author = {Carrizosa, Emilio and Ramírez-Ayerbe, Jasone and Romero Morales, Dolores},
	month = mar,
	year = {2024},
	pages = {121954},
}

@article{lara_transport-based_2024,
	title = {Transport-based counterfactual models},
	volume = {25},
	issn = {1533-7928},
	url = {http://jmlr.org/papers/v25/21-1440.html},
	number = {136},
	urldate = {2025-10-30},
	journal = {Journal of Machine Learning Research},
	author = {Lara, Lucas De and González-Sanz, Alberto and Asher, Nicholas and Risser, Laurent and Loubes, Jean-Michel},
	year = {2024},
	pages = {1--59},
}

@inproceedings{kantorovich_transfer_1942,
	title = {On the transfer of masses},
	volume = {37},
	url = {https://cir.nii.ac.jp/crid/1370565168575910170},
	urldate = {2025-10-25},
	booktitle = {Doklady {Akademii} {Nauk}},
	author = {Kantorovich, Leonid},
	year = {1942},
	pages = {199--201},
}

@article{brenier_polar_1991,
	title = {Polar factorization and monotone rearrangement of vector-valued functions},
	volume = {44},
	copyright = {Copyright © 1991 Wiley Periodicals, Inc., A Wiley Company},
	issn = {1097-0312},
	url = {https://onlinelibrary.wiley.com/doi/abs/10.1002/cpa.3160440402},
	doi = {10.1002/cpa.3160440402},
	language = {en},
	number = {4},
	urldate = {2025-10-25},
	journal = {Communications on Pure and Applied Mathematics},
	author = {Brenier, Yann},
	year = {1991},
	pages = {375--417},
}

@article{knott_optimal_1984,
	title = {On the optimal mapping of distributions},
	volume = {43},
	copyright = {http://www.springer.com/tdm},
	issn = {0022-3239, 1573-2878},
	url = {http://link.springer.com/10.1007/BF00934745},
	doi = {10.1007/BF00934745},
	language = {en},
	number = {1},
	urldate = {2025-11-03},
	journal = {Journal of Optimization Theory and Applications},
	author = {Knott, Martin and Smith, C. S.},
	year = {1984},
	pages = {39--49},
}

@inproceedings{ley_global_2022,
	title = {Global counterfactual explanations: {Investigations}, implementations and improvements},
	shorttitle = {Global counterfactual explanations},
	url = {https://openreview.net/forum?id=Btbgp0dOWZ9},
	language = {en},
	urldate = {2025-10-23},
	booktitle = {{ICLR} 2022 {Workshop} on {PAIR2Struct}: {Privacy}, {Accountability}, {Interpretability}, {Robustness}, {Reasoning} on {Structured} {Data}},
	author = {Ley, Dan and Mishra, Saumitra and Magazzeni, Daniele},
	month = mar,
	year = {2022},
}

@article{atzmueller_subgroup_2015,
	title = {Subgroup discovery},
	volume = {5},
	copyright = {© 2015 John Wiley \& Sons, Ltd.},
	issn = {1942-4795},
	url = {https://onlinelibrary.wiley.com/doi/abs/10.1002/widm.1144},
	doi = {10.1002/widm.1144},
	language = {en},
	number = {1},
	urldate = {2025-10-23},
	journal = {WIREs Data Mining and Knowledge Discovery},
	author = {Atzmueller, Martin},
	year = {2015},
	pages = {35--49},
}

@inproceedings{rawal_beyond_2020,
	title = {Beyond individualized recourse: {Interpretable} and interactive summaries of actionable recourses},
	volume = {33},
	shorttitle = {Beyond individualized recourse},
	url = {https://proceedings.neurips.cc/paper/2020/hash/8ee7730e97c67473a424ccfeff49ab20-Abstract.html},
	urldate = {2025-10-23},
	booktitle = {Advances in {Neural} {Information} {Processing} {Systems}},
	publisher = {Curran Associates, Inc.},
	author = {Rawal, Kaivalya and Lakkaraju, Himabindu},
	year = {2020},
	pages = {12187--12198},
}

@article{carrizosa_mathematical_2024,
	title = {Mathematical optimization modelling for group counterfactual explanations},
	volume = {319},
	issn = {0377-2217},
	url = {https://www.sciencedirect.com/science/article/pii/S037722172400002X},
	doi = {10.1016/j.ejor.2024.01.002},
	number = {2},
	urldate = {2025-10-23},
	journal = {European Journal of Operational Research},
	author = {Carrizosa, Emilio and Ramírez-Ayerbe, Jasone and Romero Morales, Dolores},
	month = dec,
	year = {2024},
	pages = {399--412},
}

@article{gunning_xaiexplainable_2019,
	title = {{XAI}—{Explainable} artificial intelligence},
	volume = {4},
	number = {37},
	journal = {Science Robotics},
	publisher = {American Association for the Advancement of Science},
	author = {Gunning, David and Stefik, Mark and Choi, Jaesik and Miller, Timothy and Stumpf, Simone and Yang, Guang-Zhong},
	year = {2019},
	pages = {eaay7120},
}

@article{wachter_counterfactual_2017,
	title = {Counterfactual explanations without opening the black box: {Automated} decisions and the {GDPR}},
	volume = {31},
	issn = {1556-5068},
	shorttitle = {Counterfactual {Explanations} {Without} {Opening} the {Black} {Box}},
	url = {https://www.ssrn.com/abstract=3063289},
	doi = {10.2139/ssrn.3063289},
	language = {en},
	number = {2},
	urldate = {2024-11-13},
	journal = {Harvard Journal of Law \& Technology},
	author = {Wachter, Sandra and Mittelstadt, Brent and Russell, Chris},
	year = {2017},
}
\bibliographystyle{apalike}

\newpage
$ $
\newpage

\appendix

\section{Density preservation and Lipschitz continuity of OT Maps}
In this appendix, we present original key propositions to support the feasibility and convexity of our proposal, as well as other theoretical results developed in this work. Note that the results Propositions \ref{prop:jacobian_bounds}, \ref{prop:volume_distortion}, \ref{prop:lipschitz_sigmas} and \ref{prop:loewner_bilip} are results with a very simple algebraic proof. We decided to include them for completeness.
\begin{proposition}[Convexity of the Lipschitz Constraint]
\label{prop:lipschitz_convex}
For the group counterfactual problem (Definition~\ref{def:gcfx}), the feasible region defined by the Lipschitz continuity constraint (Equation~\eqref{eq:lipschitz_carrizosa}) with $K \geq 0$ constitutes a convex set in the space of counterfactual instances $\underline{\mathbf{x}}' = \{ \mathbf{x}'^{(1)}, \dots, \mathbf{x}'^{(n)} \}$.
\end{proposition}

\begin{proof}
Let $r_{ij} = K \| \mathbf{x}^{(i)} - \mathbf{x}^{(j)} \|_2$ denote the constant upper bound derived from the input data. For any pair $(i,j)$, the constraint restricts the solution to the set:
\begin{equation}
    S_{ij} = \left\{ \underline{\mathbf{x}}' \in \mathcal{X}^n \mid \| \mathbf{x}'^{(i)} - \mathbf{x}'^{(j)} \|_2 \leq r_{ij} \right\}
\end{equation}
To establish the convexity of $S_{ij}$, consider two arbitrary feasible points $\mathbf{u}, \mathbf{v} \in S_{ij}$ and a scalar $\lambda \in [0,1]$. Let $\mathbf{z}_\lambda = \lambda \mathbf{u} + (1-\lambda)\mathbf{v}$ be their convex combination. By the linearity of vector addition and the triangle inequality, the pairwise difference satisfies:
\begin{align*}
    \| \mathbf{z}_\lambda^{(i)} - \mathbf{z}_\lambda^{(j)} \|_2 
    &= \| \lambda(\mathbf{u}^{(i)} - \mathbf{u}^{(j)}) + (1-\lambda)(\mathbf{v}^{(i)} - \mathbf{v}^{(j)}) \|_2 \\
    &\leq \lambda \| \mathbf{u}^{(i)} - \mathbf{u}^{(j)} \|_2 + (1-\lambda) \| \mathbf{v}^{(i)} - \mathbf{v}^{(j)} \|_2 \\
    &\leq \lambda r_{ij} + (1-\lambda) r_{ij} \\
    &= r_{ij}
\end{align*}
Thus, $\mathbf{z}_\lambda \in S_{ij}$, implying that $S_{ij}$ is convex. The complete feasible set is the intersection of these constraints over all pairs:
\begin{equation*}
    S = \bigcap_{i,j \in \{1,..,n\}} S_{ij}
\end{equation*}
Since the intersection of any collection of convex sets is itself convex, $S$ is a convex set.
\end{proof}

\begin{proposition}[Non-Convexity of the Bi-Lipschitz Constraint]
\label{prop:bilip_noncvx}
The bi-Lipschitz continuity constraint (Equation~\eqref{eq:bilipschitz}) for the problem in Definition~\ref{def:gcfx} defines a non-convex feasible set.
\end{proposition}

\begin{proof}
The non-convexity arises from the lower bound of the bi-Lipschitz condition. Let $d_{ij} = \|\mathbf{x}^{(i)} - \mathbf{x}^{(j)}\|_2$ be the distance between two distinct original instances ($i \neq j$), implying $d_{ij} > 0$.

The constraint for this pair requires the counterfactuals $\mathbf{x}'^{(i)}, \mathbf{x}'^{(j)}$ to satisfy:
\begin{equation}
    \frac{1}{k} d_{ij} \leq \|\mathbf{x}'^{(i)} - \mathbf{x}'^{(j)}\|_2 \leq K d_{ij}
\end{equation}
Let $r_{\min} = \frac{1}{k} d_{ij}$ and $r_{\max} = Kd_{ij}$. Since $k \geq 1$ and $d_{ij} > 0$ (we assume all points are different), we have $r_{\min} > 0$.

Consider a unit vector $\mathbf{u} \in \mathbb{R}^d$ (vector of length 1) and a scalar $r$ such that $r_{\min} \leq r \leq r_{\max}$. We construct two specific feasible solutions, $\mathbf{S}_1$ and $\mathbf{S}_2$, in the product space $\mathcal{X} \times \mathcal{X}$:
\begin{align*}
    \mathbf{S}_1 &= (\mathbf{x}'^{(i)}_1, \mathbf{x}'^{(j)}_1) = (r\mathbf{u}, \mathbf{0}) \\
    \mathbf{S}_2 &= (\mathbf{x}'^{(i)}_2, \mathbf{x}'^{(j)}_2) = (-r\mathbf{u}, \mathbf{0})
\end{align*}
Both $\mathbf{S}_1$ and $\mathbf{S}_2$ are feasible because the distance between the components in each pair is exactly $\| \pm r\mathbf{u} - \mathbf{0} \|_2 = r$, which lies within the valid range $[r_{\min}, r_{\max}]$.

Now, consider the midpoint $\mathbf{S}_m = \frac{1}{2}(\mathbf{S}_1 + \mathbf{S}_2)$:
\[
\mathbf{S}_m = \left( \frac{r\mathbf{u} - r\mathbf{u}}{2}, \frac{\mathbf{0} + \mathbf{0}}{2} \right) = (\mathbf{0}, \mathbf{0})
\]
For $\mathbf{S}_m$, the distance between the counterfactuals is $\|\mathbf{0} - \mathbf{0}\|_2 = 0$. Since $0 < r_{\min}$, $\mathbf{S}_m$ violates the lower bound of the constraint. Therefore, the feasible set contains two points whose midpoint is not in the set, proving it is non-convex.
\end{proof}

\begin{proposition}[Jacobian Bounds for OT Maps]
\label{prop:jacobian_bounds}
Let $\hat{g}: \mathcal X \to \mathcal X$ (where $\mathcal X \subset \mathbb{R}^d$) be a differentiable map that is globally $(K,k)$-bi-Lipschitz on $\mathcal X$ with $K, k \geq 1$. Then the Jacobian determinant of $\hat{g}$ satisfies the following:
\begin{equation}
    \frac{1}{k^d} \leq |\det D\hat{g}(\set{x})| \leq K^d
    \label{eq:jacobian_bounds}
\end{equation}
for all $\set{x} \in \mathcal X$.
\end{proposition}

\begin{proof}
Let $D\hat{g}(\set{x})$ denote the Jacobian matrix at $\set{x}$. The bi-Lipschitz condition implies that for any unit vector $\set{v} \in \mathbb{R}^d$:
$$ \frac{1}{k} \leq \| D\hat{g}(\set{x}) \set{v} \| \leq K. $$
These bounds on the operator norm correspond to the extremal singular values of the Jacobian. Specifically, if $\sigma_1 \ge \dots \ge \sigma_d$ are the singular values of $D\hat{g}(\set{x})$, then:
$$ \frac{1}{k} \leq \sigma_d \leq \sigma_i \leq \sigma_1 \leq K \quad \forall i \in \{1, \dots, d\}. $$
Since the absolute determinant is the product of the singular values ($|\det D\hat{g}(\set{x})| = \prod_{i=1}^d \sigma_i$), we obtain:
$$ \frac{1}{k^d} = \prod_{i=1}^d \frac{1}{k} \leq \prod_{i=1}^d \sigma_i \leq \prod_{i=1}^d K = K^d. $$
This establishes the stated bounds.
\end{proof}

\begin{proposition}[Volume Preservation in OT Counterfactual Maps]
\label{prop:volume_distortion}
Given an OT counterfactual map $\hat{g}: \mathcal X \to \mathcal X$ (with $\mathcal X \subset \mathbb R ^d$)  that is $(K,k)$-bi-Lipschitz constrained with $K, k \geq 1$, the volume $\text{Vol}(\hat{g}(\mathcal{A}))$, for any measurable set $\mathcal{A} \subset \mathcal{X}$ is bounded by:
\begin{equation}
\frac{1}{k^d}\text{Vol}(\mathcal{A}) \leq \text{Vol}(\hat{g}(\mathcal{A})) \leq K^d\text{Vol}(\mathcal{A})
\label{eq:volume_distortion}
\end{equation}
\end{proposition}

\begin{proof}
The volume of the transformed set is:
$$\text{Vol}(\hat{g}(\mathcal{A})) = \int_{\mathcal{A}} |\det D\hat{g}(\set{x})| \, d\set{x},$$
\noindent derived by change of variables from $\text{Vol}(\mathcal{A}) = \int_{\mathcal{A}} d\set{x}$.

Proposition~\ref{prop:jacobian_bounds}, specifically Equation~(\ref{eq:jacobian_bounds}), bounds the Jacobian determinant $|\det D\hat{g}(\set{x})|$. Considering $\mathcal A$ and integrating over it yields
$$\int_{\mathcal{A}} \frac{1}{k^d} \, d\set{x} \leq \int_{\mathcal{A}} |\det D\hat{g}(\set{x})| \, d\set{x} \leq \int_{\mathcal{A}} K^d \, d\set{x}.$$
Factoring out the constants from the bounds and equating the integrals to the volumes yields Equation~(\ref{eq:volume_distortion}), proving the theorem.
\end{proof}

\begin{proposition}[Optimizing the Affine Transform is a QP]
\label{prop:affine_qp}
Let $\{\set{x}^{(i)}\}_{i=1}^n\subset\mathbb R^d$ be fixed data. Let decision variables
\[
\coefsaffine\in\mathbb R^{d\times d},\qquad \interaffine\in\mathbb R^d,
\]
which can be reformulated as a $(n,1)$ shaped matrix $\set{Z}$:
\[
\set{Z}=\begin{bmatrix}\operatorname{vec}(\coefsaffine)\\[4pt] \interaffine\end{bmatrix}\in\mathbb R^{d^2+d}.
\]
Consider the objective
\[
F(\coefsaffine,\interaffine)=\sum_{i=1}^n \|\coefsaffine \set{x}^{(i)} + \interaffine - \set{x}^{(i)}\|_2^2
\]
and affine constraints of the form $W^\top (\coefsaffine \set{x}^{(i)} + \interaffine)\le r$ (e.g. logistic regression logit constraints). Then the objective is a quadratic function in \(z\) with a PSD Hessian, so the problem as defined in Definition~\ref{def:functional_gcfx} without Lipschitz or additional non-convex constraints is a convex quadratic program.
\end{proposition}

\begin{proof}
Define for each $i$ the matrix
\[
\set{M}_i=\begin{bmatrix}I_d\otimes \set{x}^{(i)\top} &\; I_d\end{bmatrix}\in\mathbb R^{d\times(d^2+d)},
\]
so that $\coefsaffine \set{x}^{(i)} + \interaffine = \set{M}_i \set{Z}$. Stack the $\set{M}_i$ vertically to form $\set{M}\in\mathbb R^{nd\times(d^2+d)}$ and stack the $\set{x}^{(i)}$ into $\tilde{\set{x}}\in\mathbb R^{nd}$. Then
\[
F(\coefsaffine,\interaffine)=\sum_{i=1}^n \|\set{M}_i \set{Z} - \set{x}^{(i)}\|_2^2 = \|\set{M} \set{Z} - \tilde{\set{x}}\|_2^2.
\]
Its expansion yields the quadratic form
\[
F(\set{Z})= \set{Z}^\top (\set{M}^\top \set{M}) \set{Z} - 2\tilde{\set{x}}^\top \set{M} \set{Z} +\tilde{\set{x}}^\top\tilde{\set{x}}.
\]
Thus the Hessian with respect to $z$ is
\[
\nabla^2_{\set{Z}} F = 2 \set{M}^\top \set{M} \succeq 0,
\]
so $F$ is a convex quadratic form in $\set{Z}$. Any affine constraints in $(\coefsaffine,\interaffine)$ (including the logistic linear constraints written above) are affine in $\set{Z}$. Therefore, optimizing $F$ subject only to affine constraints is exactly a (convex) QP.

\end{proof}

\begin{proposition}[Non-convexity of the Bi-Lipschitz constraint for arbitrary affine maps]
\label{prop:bilip_noncvx_aff}
    The bi-Lipschitz constraint forms a non-convex set for the problem of Definition~\ref{def:functional_gcfx}, assuming $\hat{g}$ is an affine transform.
\end{proposition}
\begin{proof}
    Analogously to the more generic Proposition~\ref{prop:bilip_noncvx}, we will use a counterexample, albeit more specific and relying on the simplification of the bi-Lipschitz constraint proven in Proposition~\ref{prop:lipschitz_sigmas}.

     Let $\mathcal S$ define a subset of the feasible set of the problem, containing $\mathbf{A}_1$ and $\mathbf{A}_2$:

    \begin{equation*}
        \mathbf{A}_1 = \left( \begin{matrix}
        1 & 0 \\
        0 & 1
        \end{matrix} \right)
        \qquad
        \mathbf{A}_2 = \left( \begin{matrix}
        1 & 0 \\
        0 & -1
        \end{matrix} \right)
    \end{equation*}

    Both are bi-Lipschitz compliant for any $k \geq 1$, since $\mathbf{A}_1$ does not modify the data shape and $\mathbf{A}_2$ just mirrors it for variable 2, which results in no violation of the constraint (the density and isotropy of the data is perfectly preserved). However, their midpoint $\mathbf{A}_m$ is not bi-Lipschitz compliant,
    \begin{equation*}
        \mathbf{A}_m = \left( \begin{matrix}
        1 & 0 \\
        0 & 0
        \end{matrix} \right),
    \end{equation*}
    \noindent since $\mathbf{A}_m$ is not invertible (singular matrix).
\end{proof}

\begin{proposition}[Bi-Lipschitz Bounds for Affine Maps]
\label{prop:lipschitz_sigmas}
Let $\hat{g}: \mathbb{R}^d \to \mathbb{R}^d$ be defined by $\hat{g}(\mathbf{x}) = \mathbf{A}\mathbf{x} + \mathbf{b}$. Then $\hat{g}$ is Lipschitz continuous with constant $K = \sigma_{\max}(\mathbf{A})$. Furthermore, if $\mathbf{A}$ is invertible, $\hat{g}$ is bi-Lipschitz with distortion parameter $k = 1/\sigma_{\min}(\mathbf{A})$.
\end{proposition}

\begin{proof}
For any $\mathbf{x}, \mathbf{y} \in \mathbb{R}^d$, the difference in the image is invariant to translation: $\hat{g}(\mathbf{x}) - \hat{g}(\mathbf{y}) = \mathbf{A}(\mathbf{x} - \mathbf{y})$.
The variational characterization of singular values states that for any vector $\mathbf{v} \in \mathbb{R}^d$:
\[
\sigma_{\min}(\mathbf{A}) \|\mathbf{v}\|_2 \leq \|\mathbf{A}\mathbf{v}\|_2 \leq \sigma_{\max}(\mathbf{A}) \|\mathbf{v}\|_2.
\]
Substituting $\mathbf{v} = \mathbf{x} - \mathbf{y}$ immediately yields the bi-Lipschitz inequalities:
\[
\sigma_{\min}(\mathbf{A}) \|\mathbf{x} - \mathbf{y}\|_2 \leq \|\hat{g}(\mathbf{x}) - \hat{g}(\mathbf{y})\|_2 \leq \sigma_{\max}(\mathbf{A}) \|\mathbf{x} - \mathbf{y}\|_2.
\]
Identifying $K = \sigma_{\max}(\mathbf{A})$ and the lower bound $1/k = \sigma_{\min}(\mathbf{A})$ completes the proof.
\end{proof}

\begin{proposition}[Loewner Order for Bi-Lipschitz Enforcement]
\label{prop:loewner_bilip}
Let \(\coefsaffine \in \mathbb{R}^{d\times d}\) be (symmetric) PSD, $\coefsaffine \succeq 0$, and let \(k,K \geq 1\).  
Then the following are equivalent:
\[
\frac{1}{k} I_d \preceq \coefsaffine \preceq K I_d
\ \ \Longleftrightarrow \ \
\sigma_{\max}(\coefsaffine)\le K \ \land \ \frac{1}{\sigma_{\min}(\coefsaffine)}\le k
\]
\end{proposition}

\begin{proof}
Since \(\coefsaffine\) is PSD, it admits an eigendecomposition
\(\coefsaffine = Q \Lambda Q^\top\)
with orthogonal \(Q\) and real nonnegative eigenvalues
\(\lambda_1,\dots,\lambda_d\).
The Loewner inequalities
\(\tfrac{1}{k} I_d \preceq \coefsaffine \preceq K I_d\), by definition,
are equivalent to the eigenvalue bounds
\(\tfrac{1}{k} \le \lambda_i \le K\) for all \(i\), or, equivalently,
\[ 
\lambda_{\max}(\coefsaffine)\le K \text{ and } \frac{1}{\lambda_{\min}(\coefsaffine)}\le k. \]

Another by-product of $\coefsaffine \succeq 0$ is the equality between the eigenvalues and singular values, i.e., \(\lambda_1 = \sigma_1,\dots,\lambda_n = \sigma_n\). Therefore, if we substitute $\lambda_{\max}(\coefsaffine)$ and $\lambda_{\min}(\coefsaffine)$ by $\sigma_{\max}(\coefsaffine)$ and $\sigma_{\min}(\coefsaffine)$, respectively, the proposition is proven.
\end{proof}

\begin{remark}
For a non-PSD matrix \(\coefsaffine\), Proposition~\ref{prop:loewner_bilip} is not applicable.
The bi-Lipschitz condition still corresponds to
\(\sigma_{\min}(\coefsaffine)\ge 1/k\) and \(\sigma_{\max}(\coefsaffine)\le K\),
but the equality between eigenvalues and singular values no longer holds. Consider as a counterexample the proof for Proposition~\ref{prop:bilip_noncvx_aff}, where $A_1$ and $A_2$ are symmetric, but not PSD.
\end{remark}

\section{Wasserstein Distance for Gaussian Maps}

\begin{proposition}
\label{prop:gassian_A_equiv}
Let \(\covmat_P\in\mathbb{R}^{d\times d}\) be positive definite (\(\covmat_P\succ 0\)).
Let \(\coefsaffine\in\mathbb{R}^{d\times d}\) represent the matrix of an affine transform and define the block matrix
\[
\set{M}=\begin{pmatrix}\covmat_P & \mathbf{R}\\[4pt] \mathbf{R}^\top & \covmat_Q\end{pmatrix},
\qquad\text{with }\mathbf{R}=\covmat_P \coefsaffine,
\]
where \(\covmat_Q\in\mathbb{R}^{d\times d}\) and is constrained to \(\covmat_Q\succeq 0\). Block $\set{M}$ can be understood as the Gaussian coupling of an OT plan between Gaussians.

Consider the optimization (with all other variables and additive constants fixed)
\[
\min_{\covmat_Q\succeq 0}\ \operatorname{Tr}(\covmat_Q)
\qquad\text{subject to}\qquad \set{M}\succeq 0.
\]
Then at any minimizer \(\covmat_Q^\star\) we have the equality
\[
\covmat_Q^\star \;=\; \coefsaffine^\top \covmat_P \coefsaffine.
\]
\end{proposition}

\begin{proof}
Since \(\covmat_P\succ0\) the block matrix condition \(\set{M}\succeq0\) is equivalent to the Schur complement inequality
\[
\covmat_Q \;\succeq\; \mathbf{R}^\top \covmat_P^{-1} \mathbf{R}.
\]
Substituting \(\mathbf{R}=\covmat_PA\) yields the lower bound
\[
\covmat_Q \;\succeq\; \coefsaffine^\top \covmat_P \coefsaffine.
\]
Because the trace is monotone on the cone of positive semidefinite matrices, taking traces in the matrix inequality gives
\[
\operatorname{Tr}(\covmat_Q)\;\ge\;\operatorname{Tr}\bigl(\coefsaffine^\top \covmat_P \coefsaffine\bigr).
\]
The optimization minimizes \(\operatorname{Tr}(\covmat_Q)\) over all feasible \(\covmat_Q\). Therefore the minimum is achieved when the lower bound is tight. That is, any minimizer must satisfy
\[
\covmat_Q^\star - \coefsaffine^\top\covmat_P \coefsaffine \succeq 0
\quad\text{and}\quad
\operatorname{Tr}\bigl(\covmat_Q^\star - \coefsaffine^\top\covmat_P \coefsaffine\bigr)=0.
\]
A positive semidefinite matrix with zero trace is the zero matrix. Hence
\[
\covmat_Q^\star - \coefsaffine^\top\covmat_P \coefsaffine = \mathbf{0},
\]
and therefore \(\covmat_Q^\star = \coefsaffine^\top\covmat_P \coefsaffine\), as claimed.
\end{proof}

\begin{remark}
$\mathbf{M}$ is equivalent to an affine OT map with matrix $\mathbf{A}$, but expressed in terms of a coupling (Kantorovich plan). 

The definition \(\mathbf{R} = \covmat_P \coefsaffine\) is motivated by the fact that $\set{Y}\sim\mathcal N (\mvmean_Q, \covmat_Q)$ is defined as an affine transform of $\set{X} \sim \mathcal N (\mvmean_P, \covmat_P)$, i.e., $\set{Y} = \coefsaffine \set{X} + \interaffine$.

The conditional expectation of \(\set{Y}\) given \(\set{X}\) (the optimal linear estimator in the least-squares sense) is given by
\[
\mathbb{E}[\set{Y} \mid \set{X}] = \mathbf{R}^\top \covmat_P^{-1} \set{X}.
\]
Substituting the parameterization \(\mathbf{R} = \covmat_P \coefsaffine\) yields
\[
\mathbb{E}[\set{Y} \mid \set{X}] = (\covmat_P \coefsaffine)^\top \covmat_P^{-1} X 
= \coefsaffine^\top \covmat_P \covmat_P^{-1} \set{X} 
= \coefsaffine^\top \set{X}.
\]
Thus, this parameterization explicitly identifies \(\coefsaffine^\top\) as the linear operator modeling the dependence of \(\set{Y}\) on \(\set{X}\).
\end{remark}

\begin{remark}
The proof of Proposition~\ref{prop:gassian_A_equiv} uses \(\covmat_P\succ0\). If \(\covmat_P \succeq 0\) (which might be singular), the Schur complement argument must be replaced by a limiting or regularized argument, and equality need not hold in the same form. In numerical practice, equality holds up to solver tolerances.
\end{remark}

\begin{proposition}
\label{prop:wass_norm_scaled}
Let $P= N(\mvmean_P,\covmat_P)$ and $ Q = \mathcal N(\mvmean_Q,\covmat_Q)$ with $\covmat_Q=s\covmat_P$ for some scalar $s>0$.  
Then the squared $W_2$ distance satisfies
\[
W_2(P,Q)^2=\|\mvmean_P-\mvmean_Q\|^2+(\sqrt s-1)^2\operatorname{Tr}(\covmat_P).
\]
\end{proposition}

\begin{proof}
The formula above is derived from the well-known closed-form expression of the $W_2$ distance between multivariate normal distributions (Equation~(\ref{eq:wass_normal})).

We now simplify Equation~(\ref{eq:wass_normal}) for the special case $\covmat_Q = s\covmat_P$, with $s>0$.

First, we compute the second term:
\[
\covmat_Q^{1/2}\covmat_P\covmat_Q^{1/2}
= (\sqrt{s}\,\covmat_P^{1/2})\,\covmat_P\,(\sqrt{s}\,\covmat_P^{1/2})
= s\,\covmat_P^2.
\]
Taking the matrix square root yields
\[
(\covmat_Q^{1/2}\covmat_P\covmat_Q^{1/2})^{1/2}
= (s\,\covmat_P^2)^{1/2}
= \sqrt{s}\,\covmat_P.
\]

Substituting this result into the general expression:
\begin{align*}
& W_2(P,Q)^2 =  \\
&= \|\mvmean_P - \mvmean_Q\|^2 
 + \operatorname{Tr}\!\big(\covmat_P + s\covmat_P - 2\sqrt{s}\,\covmat_P\big)
\end{align*}

We factor out $\covmat_P$ and use linearity of the trace:
\[
\operatorname{Tr}\!\big(\covmat_P + s\covmat_P - 2\sqrt{s}\,\covmat_P\big)
= (1 + s - 2\sqrt{s})\,\operatorname{Tr}(\covmat_P).
\]
Finally, the scalar coefficient is simplified, yielding
\[
W_2(P,Q)^2 = \|\mvmean_P - \mvmean_Q\|^2 + (\sqrt{s}-1)^2\,\operatorname{Tr}(\covmat_P),
\]
which proves the theorem.

\end{proof}

\subsection{Probabilistic OT with Gaussian Mixture Models}
\label{sec:probabilistic_gcfx}

In scenarios where the data distribution is multi-modal and/or not absolutely continuous~\citep{brenier_polar_1991}, affine-based maps (and Gaussian counterfactuals by extension) may fail not yield an optimal solution. To address this, we present a formulation using GMMs, which serve as universal approximators capable of modeling arbitrary densities while retaining the computationally tractable properties of Gaussian OT (Equation~(\ref{eq:wass_normal})).

We assume the source group distribution $P$ is parameterized as a GMM with $m>1$ components. Crucially, we seek a counterfactual distribution $Q$ that preserves this mixture structure. The distributions are defined as:
\begin{equation}
    P = \sum_{j=1}^m w_j \mathcal{N}(\mvmean_{Pj}, \covmat_{Pj}), \quad Q = \sum_{j=1}^m w_j \mathcal{N}(\mvmean_{Qj}, \covmat_{Qj})
    \label{eq:gmm}
\end{equation}
\noindent where $\sum_j^m w_j = 1$.

Instead of solving for a single transport map, we decompose the problem into $m$ sub-problems. We assume a component-wise coupling where the $j$-th Gaussian of $P$ is mapped to the $j$-th Gaussian of $Q$ via a specific affine transformation $A_j\set{x} + b_j$. Consequently, minimizing the total transport cost reduces to applying Definition~\ref{def:functional_gcfx_norm} individually to each pair of components:
\begin{equation}
    W_2(P, Q)^2 \approx \sum_{j=1}^m w_j \cdot W_2(P_j, Q_j)^2
\end{equation}
\noindent where $P_j = \mathcal{N}(\mvmean_{Pj}, \covmat_{Pj})$ and $Q_j = \mathcal{N}(\mvmean_{Qj}, \covmat_{Qj})$ refer to $j$-th component of the GMM.

This implicitly defines $m$ distinct affine maps $\{\coefsaffine_j, \interaffine_j\}_{j=1}^m$ and optimizes the parameters of the counterfactual components $\mathcal{N}(\mvmean_{Qj}, \covmat_{Qj})$.

However, the $m$ problems cannot be solved independently, as the bi-Lipschitz continuity should be applied to the whole transportation plan. This is a non-trivial task, as the global map is piecewise or probabilistic. We propose a relaxation that enforces robustness at two levels: \textit{local} and \textit{global}.

First, we strictly enforce the bi-Lipschitz constraint on each individual map $\coefsaffine_j$ to ensure local stability within each cluster. Second, to maintain global coherence, we constrain the expectation of the projection distance:
\begin{equation}
\begin{split}
         & \tfrac{1}{k} \|\set{x}^{(i)} - \set{x}^{(j)}\|_2 \leq \mathbb E[\|\set{x}'^{(i)}- \set{x}'^{(j)}\|_2] \leq K \|\set{x}^{(i)} -\set{x}^{(j)}\|_2, \\
     & \forall i,j \in \{1..n\}, \ i\neq j
    \label{eq:bilipschitz_relax}
\end{split}
\end{equation}
The expectation in Equation~(\ref{eq:bilipschitz_relax}) corresponds to an affine map $(\sum_j^m w_j \coefsaffine_j) \set{x} +(w_j \interaffine_j)$. 

\section{Experimentation Appendix}

\subsection{Model Selection and Training}
\label{app:model_training}
We use L2 regularized logistic regression, as it is a well-known and simple linear classifier. 
Although it will often perform worse than other models, such as gradient boosting trees, maximizing predictive performance or finding the optimal model for a given data set is not the focus of this work; rather, it is to compare group counterfactual techniques for a user-given model.
To create a more robust model, we use an L2-regularized version. 
We run a simple grid search on the penalty hyperparameter $\gamma$, on log-scale, in the interval \([10^{-2}, 10^2]\), including 0 (no penalty, standard logistic regression), 
and select the best value via 10-fold cross-validation and minimal cross-entropy (Table~\ref{tab:datasets} displays the results).

\subsection{Performance Metrics for Group Counterfactuals}
\label{app:metrics}
The metrics considered in this work for evaluating counterfactuals and their explanations are as follows. 
Except for the first one ($W_2$ distance), the metrics are originally proposed in this work or extensions of known counterfactual metrics to the group case (as in validity).
\begin{enumerate}
    \item Empirical squared $W_2$ distance. Given an input group $\underline{\set{x}} = \{\set{x}^{(i)}\}_{i=0}^n$ and its respective counterfactual group $\underline{\set{x}'} = \{\set{x}'^{(i)}\}_{i=0}^n$, the squared $W_2$ distance is defined as
    \begin{equation}
        \frac{1}{n}\sum^{n}_{i=1} || \set{x}^{(i)} - g(\set{x}^{(i)}) ||_2^2, \ \forall i \in \{1,..,n\},
    \end{equation}
    \noindent which is equivalent to Equation~(\ref{eq:wass}) for finite samples (rather than known distributions) and under the assumption that $\set{x}'^{(i)}$ is the corresponding point of $\set{x}^{(i)}$.
    \item Empirical (bi-)Lipschitz upper bound (right term of Equation~(\ref{eq:lipschitz_carrizosa}) or Equation~(\ref{eq:bilipschitz}), which measures the maximum dispersion of the map $\hat{g}$. It is defined as the minimum value for $K$ given an input group $\underline{\set{x}}$ and a counterfactual map $\hat{g}$. Formally, 
    \begin{equation}
        \hat{K} = \min_{i,j} \frac{|| \set{x}^{(i)} - \set{x}^{(j)} ||_2}{|| \hat{g}(\set{x})^{(i)} - \hat{g}(\set{x})^{(j)} ||_2}.
        \label{eq:emp_upper_bound}
    \end{equation}
    \item Empirical bi-Lipschitz lower bound (left term of Equation~(\ref{eq:bilipschitz})), which measures the minimum compression of a map $\hat{g}$. It is defined as the maximum value for $\frac{1}{k}$ (or equivalently, the minimum $k$) given an input group $\underline{\set{x}}$ and a counterfactual map $\hat{g}$. Formally, 
    \begin{equation}
        \hat{k} = \min_{i,j} \frac{|| \hat{g}(\set{x})^{(i)} - \hat{g}(\set x)^{(j)} ||_2}{|| \set{x}^{(i)} - \set{x}^{(j)} ||_2}.
        \label{eq:emp_lower_bound}
    \end{equation}
    \item The total distortion, a summary measure that combines the empirical lower and upper bi-Lipschitz bounds. The distortion accounts for both data dispersion and compression. It is defined as:
    \begin{equation}
        \hat{D} = 1 - \frac{1}{max\{\hat{k},\hat{K}\}}
    \label{eq:distortion}
    \end{equation}
    Inverting and subtracting $1$ keeps the scale in the range $(0,1)$. 
    Smaller values indicate better bi-Lipschitz compliance, i.e., less distortion.
    \item Counterfactual validity, i.e., if the obtained counterfactuals belong to the defined counterfactual class $c'$. Validity for group counterfactuals is considered as presented in Definition~\ref{def:validity}.

    \begin{definition}[Group counterfactual validity]
    \label{def:validity}
        Consider the OT group counterfactual problem, as formalized in Definition~\ref{def:functional_gcfx}. The validity $\hat{V}$ of the functional group counterfactual $\hat{g}(\cdot)$ w.r.t a group of instances $\underline{\set{x}'} = \{\set{x}'^{(i)}\}_{i=1}^n$, the target label $c'$, score function $\hat{s}(\cdot, \cdot)$ and score threshold $\alpha$, is defined as:

        \begin{equation}
            \hat{V} = \frac{1}{|\underline{\set{x}'}|}\sum_{i=1}^n \mathbb I \left(\hat{s}(\hat{g}(\set{x}^{(i)}),c') > \alpha \right).
        \end{equation}
    \end{definition}

    \begin{remark}
        In Definition~\ref{def:validity}, the validity is always 1 (maximum value) if the group $\underline{\set{x}}$ has been used to derive the functional counterfactual $\hat{g}$. The definition only makes sense when applied to unseen data points during training.
    \end{remark}
\end{enumerate}


\subsection{Unbiased Estimation of Metrics}
\label{app:cross_val}
As a first crude possibility, one could find $\hat{g}$ for a given group $\underline{\set{x}}$, ``predict'' the counterfactual group $\underline{\set{x}'} = \hat{g}(\underline{\set{x}})$ and measure a selected metric using $\underline{\set{x}}$ and $\underline{\set{x}'}$, and compare it with that obtained using the baselines.

However, the previous approach defeats the purpose of finding OT maps $\hat{g}$ that generalize to new instances. 
Hence, we estimate our metrics under 10-fold cross-validated, i.e., find $\hat{g}$ with 90\% of the group $\underline{\set{x}}$, ``predict'' the counterfactual group for the remaining 10\% of the data, and obtain the metrics for it, averaging over the 10 folds.

The results presented in the figures throughout the manuscript always refer to validation values for the metrics using the described procedure (except for the baselines, where generalization is not possible, so the metric is computed on the full train dataset).

\subsection{Sensitivity to $K$}
\label{app:k_sensitivity}
In the main text, we claim that the $W_2$ distance decreases with increasing $K$, i.e., it loosely enforces the bi-Lipschitz constraint. 
This is also visible in Section~\ref{sec:bn} through the Bayesian network surrogate.

\begin{figure}[t]
    \includegraphics[width=.47\textwidth]{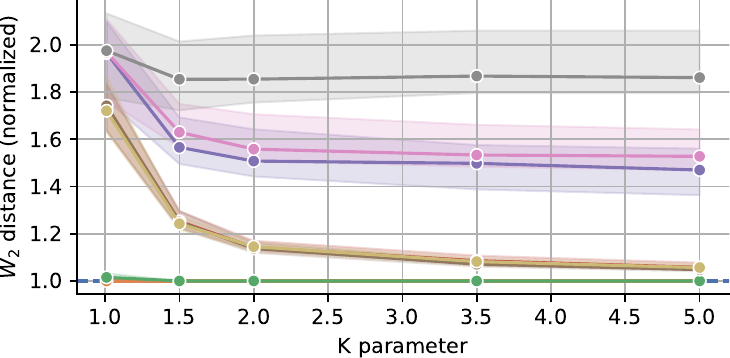}
    \caption{Normalized $W_2$ distance per $K$.}
    \label{fig:lineplot_wass}
\end{figure}

In this appendix, we provide further proof through Figure~\ref{fig:lineplot_wass}. 
The $W_2$ is normalized by dividing by the value of the independent algorithm (first baseline, Table~\ref{tab:model_comparison}). 
Thus, values represent the $W_2$ distance ratio with respect to the independent algorithm.

It is visible how the methods converge to the value of the independent algorithm as $K$ increases. 
The $W_2$ distance for the group counterfactual found by the dense transforms is less than 1.1 times that of the baseline for $K=5$. 
For strict enforcements ($K=1.01$) it can seem that our proposal underperforms. 
However, we compare our methods against two baselines that do not enforce the bi-Lipschitz constraint. Furthermore, the line for the group algorithm w/ bi-Lipschitz constraint shows a very optimistic case for this transform, as we remove from the plot the experiments in which the group algorithm w/ bi-Lipschitz constraint does not converge (35\% of experiments).

\subsection{Multiobjective Group Counterfactual Optimization}
\label{sec:case_heur}
For this experiment, we still use the learned logistic regression, but assume its components (coefficients and gradients) cannot be accessed, hence simulating scenarios where the model is a black box with unknown parameters. 

The distortion (bi-Lipschitz parameter $K$) is removed as a constraint but added as an additional objective. 
The two objectives to be considered are then the squared $W_2$ distance and the distortion $\hat{D}$ as defined in Appendix~(\ref{app:metrics}).

We decide to test our proposals using a gradient-free metaheuristic, as it avoids any assumptions about the model. 
Specifically, we rely on NSGA-II (which is a standard solution and which has been successfully used in other multiobjective counterfactual works \citep{dandl_multi-objective_2020}), but other options can also be considered. 

The range of objectives varies significantly across datasets and groups. 
To ensure fair comparability, we apply max-min normalization to rescale each objective to the interval $[0, 1]$ for each experiment. 
We then compute the hypervolume for these normalized objectives relative to the reference point $(0, 0)$ (i.e., lower hypervolume indicates better results). 
As in previous experiments, we use performance profiles to aggregate and analyze the results across the entire test suite.

\begin{figure}[t]
        \includegraphics[width=.47\textwidth]{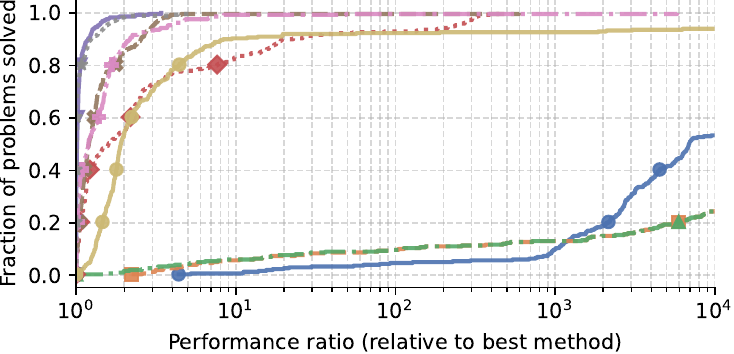}
        \caption{Hypervolume ratio.}
        \label{fig:heur_pp}
    \end{figure}

The results are displayed in Figure~\ref{fig:heur_pp}. 
All our proposals greatly exceed the baselines. 
Sparser maps, specifically the diagonal affine transform and the scaled Gaussian map, are better options, with performance ratios of approximately 1.01 and 1.4 (respectively) across 80\% of the experiments. 
The densest are those that perform the worst, with the PSD $A$ affine transform and the 3-GMM being surpassed by any other alternative in almost all experiments. 
The Gaussian and commutative Gaussian maps are an intermediate option, with around 1.6 the performance ratio for 80\% of the experiments and no statistically significant differences between them. 
It is also noticeable that the Gaussian maps perform better than their affine counterparts.

\subsection{Runtime Comparison}
\label{app:times}
We consider the time comparison a ``secondary'' result, as group counterfactuals are currently not well-studied, and we mainly care about the correctness of the results; also, it is highly unlikely that any implementation of any considered method is fully optimized. 

\begin{figure}[ht]
    \centering
    \includegraphics[width=0.95\linewidth]{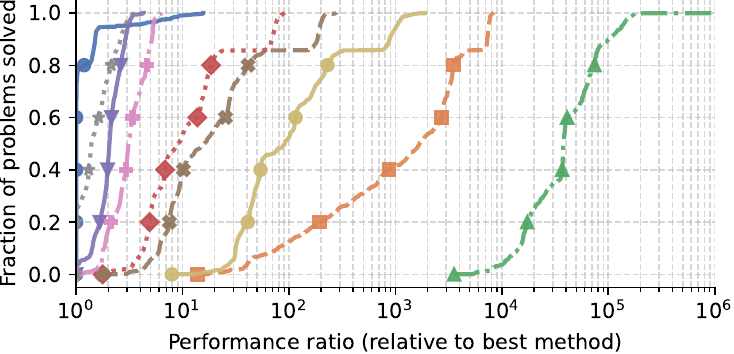}
    \caption{Time performance profile}
    \label{fig:math_time}
\end{figure}

The results are presented in Figure~\ref{fig:math_time}. 
The trend indicates that algorithms with fewer parameters and restrictions are faster. 
In 80\% of cases, the fastest algorithm is the Wachter algorithm (no Lipschitz constraints), with the scaled Gaussian being the fastest in the remaining 20\% of scenarios. 
Our sparse proposals are the fastest, performing only between 1.1 and 1.4 times slower than the fastest proposal in 80\% of the experiments. 
The dense proposals are slower, with the 3-GMM performing 105 times slower in 80\% of the experiments. 
However, the speed-up is still substantial when comparing our proposals to the baseline group counterfactual algorithm with both Lipschitz and bi-Lipschitz constraints. 
The former is around 1250 times slower than the fastest proposal, and adding the bi-Lipschitz constraint results in the baseline being almost 60000 times slower (in 80\% of the experiments). 
Compared with this figure, we can conclude that all of our proposals perform significantly faster.

\begin{figure}[ht]
    \centering
    \includegraphics[width=0.95\linewidth]{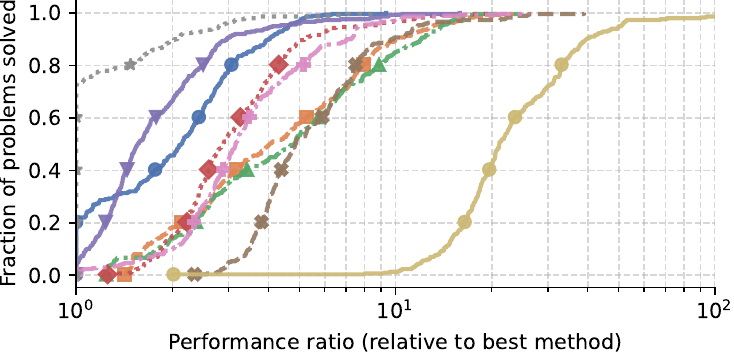}
    \caption{Time performance profile (NSGAII for multiobjective optimization).}
    \label{fig:heur_time}
\end{figure}

Figure~\ref{fig:heur_time} shows the runtime performance profile for the multiobjective experiment (Appendix ~\ref{sec:case_heur}). 
It shows that the time difference between algorithms is less extreme. 
The same trend as before can be immediately spotted: Algorithms with fewer parameters are faster. 

\subsection{Gaussian Bayesian Networks as Surrogate Models}
\label{app_bn}
\subsubsection{Overview of Gaussian Bayesian Networks}
We use Gaussian Bayesian Networks (GBNs) to model dependencies among variables. 
A GBN represents the joint probability distribution as a multivariate normal, decomposing it according to a directed acyclic graph (DAG). 
Each variable $X_i$ is modeled as a linear Gaussian function of its parents $Pa(X_i)$ in the network:

\begin{equation}
    f_i(X_i \mid Pa(X_i)) \sim \mathcal{N}\left(\mvmean_i + \sum_{Y_j \in Pa(X_i)} \beta_{ij}y_j, \sigma_i^2\right)
\end{equation}

\noindent where $\mvmean_i$ is the intercept term, $\beta_{ij}$ represents the regression coefficients (edge weights) associated with parent $Y_j$, and $\sigma_i^2$ is the variance term (independent of the parents).

This formulation allows the GBN to serve as an efficient surrogate model, capturing linear dependencies and uncertainty between the original input features and target variables.

\subsubsection{Learning surrogate Bayesian networks}
The structure (DAG) of the Bayesian networks used in this work is learned using the PC algorithm \citep{spirtes_causation_2000}, which is able to discover causal relations under certain assumptions, such as causal sufficiency. As such, the arcs derived in this work can have a certain causal sense and are not merely associative. The parameters are learned through maximum likelihood estimation.

To ensure that the surrogate model adheres to logical consistency, we impose structural constraints. Specifically, we prohibit the following arcs:

\begin{itemize}
    \item From counterfactual to original variables.
    \item Between counterfactual variables (to avoid confounding effects).
    \item From $K$ to the original variables.
\end{itemize}

\subsection{Reproducibility and Code Availability}
The datasets are publicly available on OpenML. 
A summary of the data used can be seen in Table~\ref{tab:datasets}, 
and the grid search results for logistic regression are presented in Table~\ref{tab:lr_train}.

\begin{table}[ht]
\centering
\caption{Dataset summary. The number of features and instances is shown}
\label{tab:datasets}
\begin{tabular}{l|ccr}
\toprule
\bfseries Name & \bfseries \makecell{OpenML \\ ID} & \bfseries \makecell{\# \\ Feat.} & \bfseries \makecell{\# \\ Inst.} \\
\midrule
credit & 44089 & 10 & 16714 \\
california & 44090 & 8 & 20634 \\
wine & 44091 & 11 & 2554 \\
electricity & 44120 & 7 & 38474 \\
covertype & 44121 & 10 & 566602 \\
pol & 44122 & 26 & 10082 \\
house\_16H & 44123 & 16 & 13488 \\
kdd\_ipums\_la\_97-small & 44124 & 20 & 5188 \\
MagicTelescope & 44125 & 10 & 13376 \\
bank-marketing & 44126 & 7 & 10578 \\
phoneme & 44127 & 5 & 3172 \\
MiniBooNE & 44128 & 50 & 72998 \\
Higgs & 44129 & 24 & 940160 \\
eye\_movements & 44130 & 20 & 7608 \\
jannis & 44131 & 54 & 57580 \\
\bottomrule
\end{tabular}
\end{table}

\begin{table}[ht]
\centering
\caption{Logistic regression models used. Best L2 penalty multiplier and cross entropy (CE) obtained. Our chance baseline is $p=0.5$, which means that any model reporting a CE below $-ln(0.5) \approx 0.69$ is above chance.}
\label{tab:lr_train}
\begin{tabular}{l|cc}
\toprule
\bfseries Name & \bfseries L2 & \bfseries CE \\
\midrule
credit & 0 & 0.57 \\
california & 0 & 0.39 \\
wine & 0.60 & 0.53 \\
electricity & 0.60 & 0.52 \\
covertype & 2.15 & 0.65 \\
pol & 0.01 & 0.32 \\
house\_16H & 0.17 & 0.44 \\
kdd\_ipums\_la\_97-small & 0 & 0.32 \\
MagicTelescope & 2.15 & 0.49 \\
bank-marketing & 2.15 & 0.52 \\
phoneme & 2.15 & 0.52 \\
MiniBooNE & 0 & 0.26 \\
Higgs & 2.15 & 0.64 \\
eye\_movements & 27.8 & 0.68 \\
jannis & 0.17 & 0.53 \\
\bottomrule
\end{tabular}
\end{table}

The code is developed in Python and structured as a simple and extensible library that allows users to implement new OT maps and optimizers\footnote{Code repository available at \url{https://github.com/Enrique-Val/ot-group-counterfactual}}.

For all the methods shown in Table~\ref{tab:model_comparison} (except the 3rd entry, Group w/ bi-Lipchitz) the Mosek optimizer \citep{mosek_aps_mosek_2025} is used. Mosek is a state-of-the-art optimizer for convex problems that offers a good handling of SDP and a very abstract interface through the Python library \texttt{cvxpy} \citep{diamond_cvxpy_2016, agrawal_rewriting_2017}. Furthermore, the academic license has no restrictions on the amount of instances run in parallel.

The method ``Group w/ bi-Lipschitz'' is non-convex, hence it cannot be handled with Mosek. Alternatively, we use the Gurobi optimizer \citep{gurobi_optimization_llc_gurobi_2024}, one of the most powerful optimizers for non-convex problems. We use it in Python through the \texttt{pyomo} library \citep{hart_pyomo_2011,bynum_pyomo_2021}. The parameters are set to default except for\footnote{For a detailed documentation of the parameters, refer to \url{https://docs.gurobi.com/projects/optimizer/en/current/reference/parameters.html}}:
\begin{itemize}
    \item \texttt{NonConvex = 2}
    \item \texttt{TimeLimit = 1800}. Time limit in seconds.
    \item \texttt{NoRelHeurTime = 100}. Invoke a heuristic prior to solving, with a time limit of 100 seconds.
    \item \texttt{MIPFocus = 1}. More focus into finding feasible solutions.
\end{itemize}

For the multiobjective optimization experiment, we use the NSGAII algorithm and the \texttt{pymoo} library \citep{blank_pymoo_2020} to code the experiments. The parameters of the genetic algorithm are left to their default values and the main operators selected:
\begin{itemize}
    \item Simulated binary crossover.
    \item Polynomial mutation.
    \item Tournament selection.
\end{itemize}

\section{Statistical significant of the results}
\label{app:stats}
We use critical difference diagrams (CDDs) to visualize the result of the Friedman test (with Bergmann-Hommel post-hoc procedure). The diagram plots the average rank of each method across all datasets on a linear axis, where a lower rank indicates better performance. In the visualization, groups of algorithms whose average ranks whose difference is not deemed significant by our test (assuming a p-value of 0.05) are connected by a thick horizontal line, indicating that their performance is statistically indistinguishable. In contrast, algorithms not connected by a line are considered significantly different.

\begin{figure}[htbp]
    \begin{subfigure}[h]{0.48 \textwidth}
        \includegraphics[width = \textwidth]{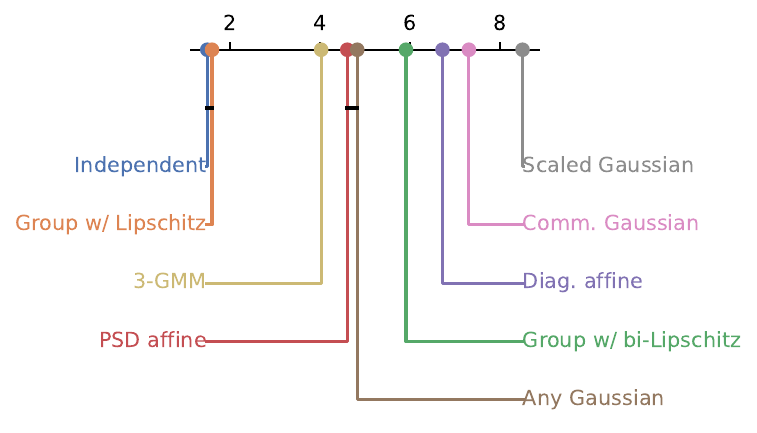}
        \caption{CDD for $K=1.01$} 
    \end{subfigure}
    \begin{subfigure}[h]{0.48 \textwidth}
        \includegraphics[width = \textwidth]{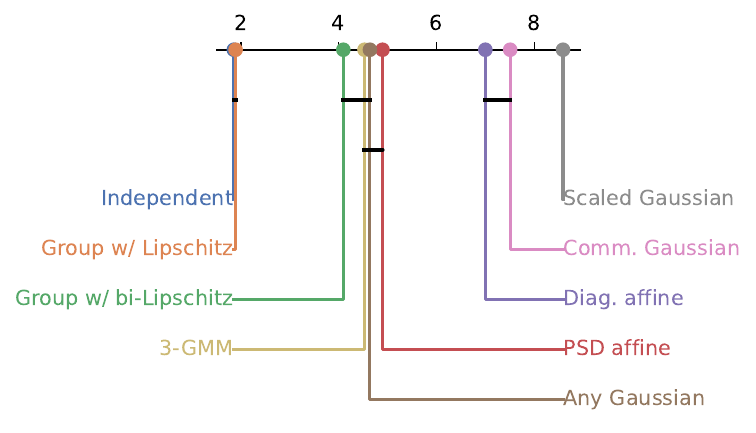}
        \caption{CDD for $K=5$} 
    \end{subfigure}
    \caption{Squared $W_2$ CDD for different $K$ values.}
    \label{fig:cdd_wass}
\end{figure}

\paragraph{Squared $W_2$ distance}
Figure~\ref{fig:cdd_wass} shows the CDD for the squared $W_2$ for the two most extreme values of $K$ tested. It is visible how the performance for the bi-Lipschitz-constrained group counterfactual algorithm degenerates for higher $K$, mainly due to non-convergence, being surpassed for 3 of our proposals

\paragraph{Empirical Lipschitz (upper) bound}
Figure~\ref{fig:cdd_upper} shows the CDD for the empirical value of $\hat{K}$ (as defined in Appendix~\ref{app:metrics}) for a parameter value of $K=1.01$, i.e., strict enforcement. It can be seen that the Wachter algorithm, which never actually enforces this constraint, is the algorithm with the lowest (stricter) upper bound, with no significant difference from the same algorithm applying the Lipschitz constraint. This proves that the Lipschitz upper bound is, in many scenarios, not necessary to explicitly enforce.

\begin{figure}[htbp]
    \centering
    \includegraphics[width=0.98\linewidth]{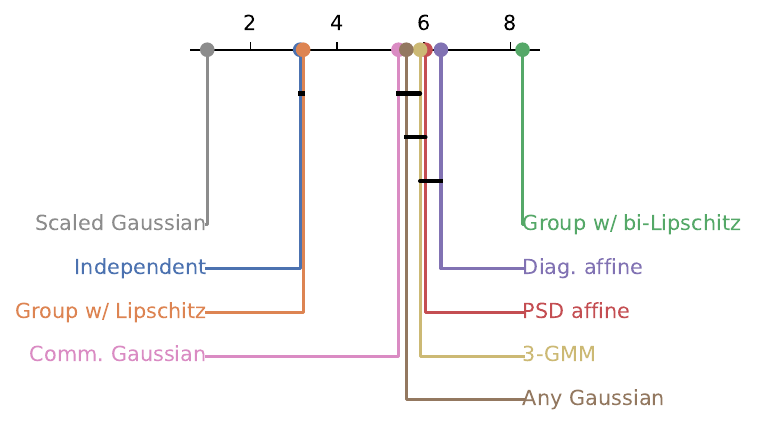}
    \caption{Empirical Lipschitz bound CDD for $K=1.01$.}
    \label{fig:cdd_upper}
\end{figure}

\begin{figure}[htbp]
    \begin{subfigure}[h]{0.48 \textwidth}
        \includegraphics[width = \textwidth]{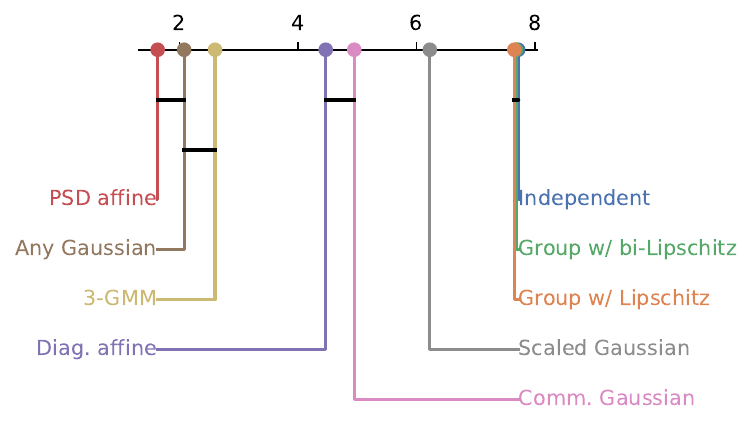}
        \caption{CDD for $K=1.01$} 
    \end{subfigure}
    \begin{subfigure}[h]{0.48 \textwidth}
        \includegraphics[width = \textwidth]{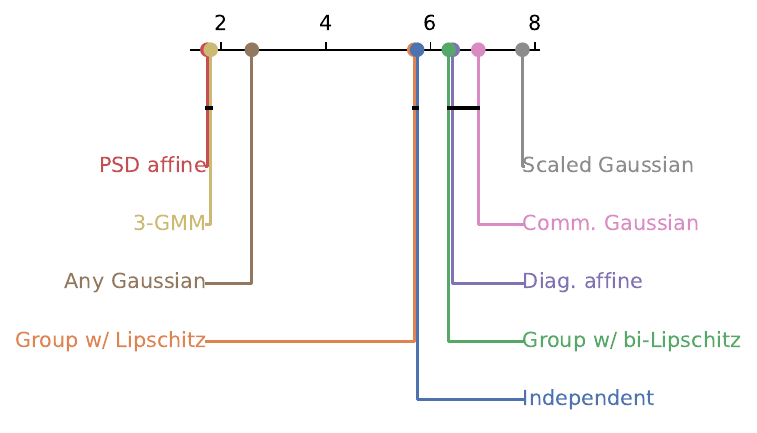}
        \caption{CDD for $K=5$} 
    \end{subfigure}
    \caption{Empirical bi-Lipschitz lower bound CDD for different $K$ values.}
    \label{fig:cdd_lower}
\end{figure}

\paragraph{Empirical bi-Lipschitz lower bound}
Figure~\ref{fig:cdd_lower} shows the CDD for the empirical bi-Lipschitz lower bound (Equation~(\ref{eq:emp_lower_bound})). Better values imply tighter bounds (i.e., low values for $\hat{k})$. For low $K$ values ($K=1.01$), our proposals offer a tighter lower bound than the baseline. While it is obvious that the independent and group w/ Lipschitz algorithm will perform worse (as they do not enforce the lower bound), we theorize that the group algorithm w/ bi-Lipschitz constraint is also surpassed because it fails to converge in many experiments. 

Regarding our proposals, denser methods offer tighter bounds. For $K=5$, the difference in performance between the sparser methods and the baseline becomes non-significant and the scaled Gaussian becomes the method with worse bounds.

\begin{figure}[htbp]
    \begin{subfigure}[h]{0.48 \textwidth}
        \includegraphics[width = \textwidth]{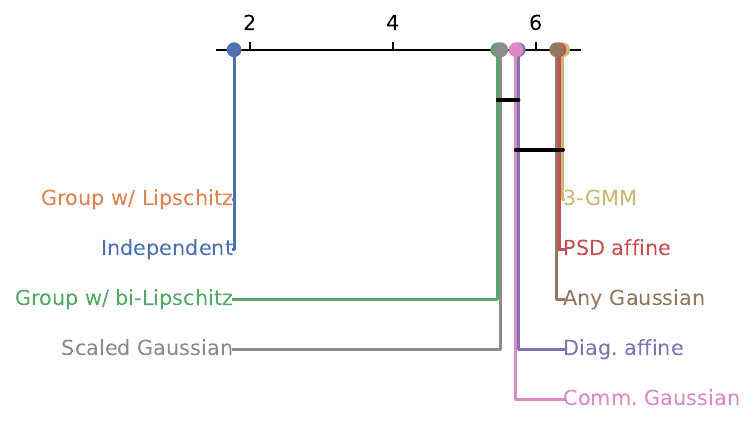}
        \caption{CDD for $K=1.01$} 
    \end{subfigure}
    \begin{subfigure}[h]{0.48 \textwidth}
        \includegraphics[width = \textwidth]{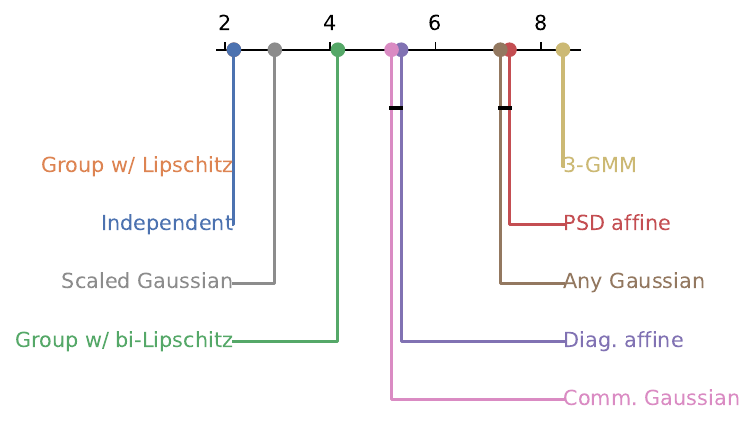}
        \caption{CDD for $K=5$} 
    \end{subfigure}
    \caption{Validity CDD for different $K$ values.}
    \label{fig:cdd_validity}
\end{figure}

\paragraph{Validity}
In Figure~\ref{fig:cdd_validity} it can be seen how the baselines ``independent'' and ``group w/ Lipschitz'' are the best performing proposals for both $K$ values. However, this is due to the fact that these proposals do not generalize and the validity needs to be estimated with train data, which, by theoretical definition, results in perfect validity. The bi-Lipschitz baseline does not achieve a similar performance in terms of validity due to non-convergence difficulties, which is especially noticeable for $K=1.01$.

Regarding our proposals, for both $K$ values sparser options offer a higher validity. However, most of these differences become significant only for high values of $K$. Although the ranking (especially for $K=1.01$) may suggest that our algorithms propose counterfactuals that are outright invalid, this is because the critical difference diagram does not take into account magnitudes between rankings. Figure~\ref{fig:validity} still proves that all our proposals generate counterfactuals with high validity.

\begin{figure}
    \centering
    \includegraphics[width=0.98\linewidth]{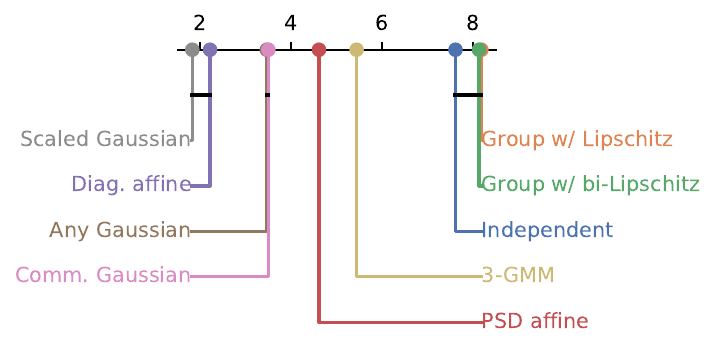}
    \caption{Hypervolume CDD}
    \label{fig:cdd_hypervolume}
\end{figure}

\paragraph{Pareto hypervolume (multiobjective optimization)}
Figure~\ref{fig:cdd_hypervolume} confirms that the sparsest transforms are the top performing ones. The baselines are classified as the worst solutions, with no significant difference between them. It is also noticeable that Gaussian maps tend to offer significantly better solutions than their affine counterparts.

\end{document}